\documentclass[letterpaper, 10 pt, conference]{ieeeconf}

\usepackage{times}                 
\usepackage[T1]{fontenc}
\usepackage[utf8]{inputenc}

\usepackage{graphicx}
\usepackage{amsmath,amssymb,mathtools}
\usepackage{bm}
\usepackage{booktabs}
\usepackage{array}
\usepackage{multirow}
\usepackage[table]{xcolor}
\usepackage{wrapfig}
\usepackage[caption=false,font=footnotesize]{subfig}  
\makeatletter
\let\labelindent\relax   
\makeatother
\usepackage{enumitem}
\usepackage{listings}
\usepackage{textcomp}

\usepackage{cite}

\usepackage[hidelinks]{hyperref}
\usepackage[nameinlink,capitalise]{cleveref}

\def\vtheta{{\bm{\theta}}}
\def\vepsilon{{\bm{\epsilon}}}
\def\vphi{{\bm{\phi}}}

\def\vq{{\bm{q}}}

\def\vs{{\bm{s}}}

\def\vx{{\bm{x}}}
\def\vy{{\bm{y}}}

\newtheorem{theorem}{Theorem}
\newtheorem{lemma}{Lemma}

\DeclarePairedDelimiterX{\infdivx}[2]{(}{)}{#1\;\delimsize\|\;#2}
\newcommand{\infdiv}{D_\textrm{KL}\infdivx}
\newcommand\scalemath[2]{\scalebox{#1}{\mbox{\ensuremath{\displaystyle #2}}}}

\title{\LARGE \bf Factorizing Diffusion Policies for Observation Modality Prioritization}

\IEEEoverridecommandlockouts
\author{Omkar Patil$^{1}$, Prabin Kumar Rath$^{1}$, Kartikay Pangaonkar$^{1}$, Eric Rosen$^{}$, Nakul Gopalan$^{1}$%
\thanks{$^{1}$School of Computing and Augmented Intelligence, Arizona State Uni.}%
\thanks{Email: {\tt\small opatil3@asu.edu}}%
}  

\hypersetup{
  pdftitle={Factorizing Diffusion Policies for Observation Modality Prioritization},
  pdfauthor={Omkar Patil, Prabin Rath, Kartikay Pangaonkar, Eric Rosen, Nakul Gopalan}
  pdfsubject={Robotics},
  pdfkeywords={Multimodal learning, Diffusion Models, Robot Learning}
}

\begin{document}
\maketitle

\begin{figure*}[!t]
  \centering
  \includegraphics[width=0.9\textwidth]{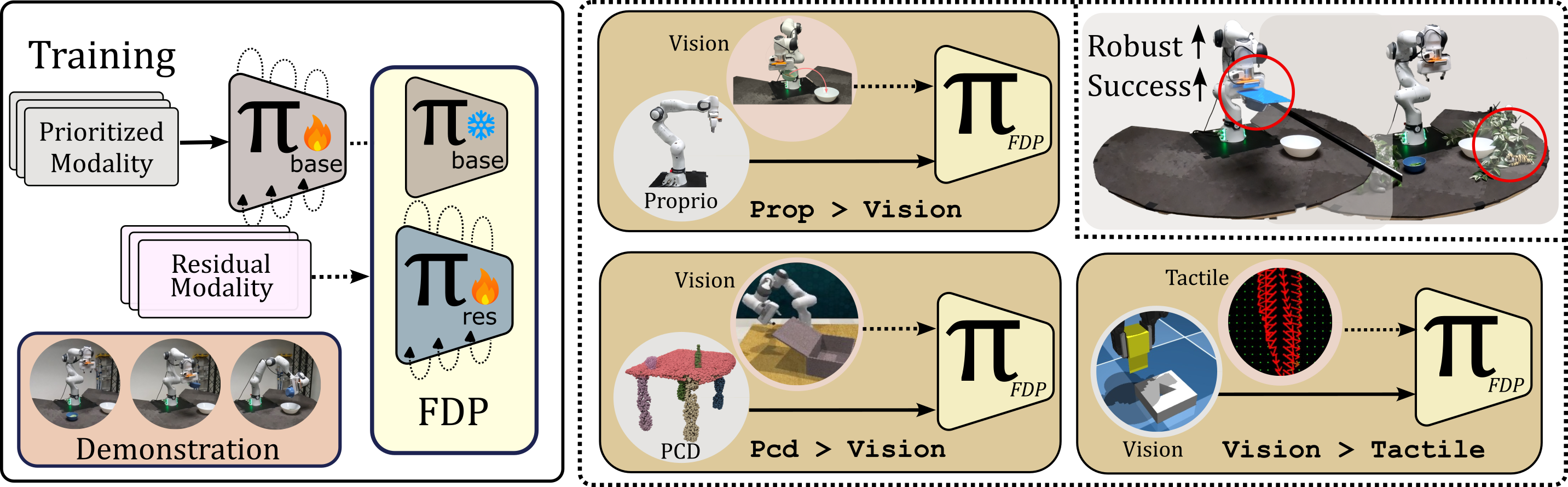}
  \caption{Policy learning using FDP with different prioritization orders. In FDP, we train a base and a residual policy by prioritizing over different observation modalities for the same task. We demonstrate this approach with different combinations of observation modalities such as \texttt{prop>vision}, and \texttt{vision>tactile} among others. FDP results in more performant policies ($15\%\uparrow$) that are robust to distractors and camera occlusions ($40\%\uparrow$).}
  \label{fig:poster}
\end{figure*}

\begin{abstract}
Diffusion models have been extensively leveraged for learning robot skills from demonstrations. These policies are conditioned on several observational modalities such as proprioception, vision and tactile. However, observational modalities have varying levels of influence for different tasks that diffusion polices fail to capture. In this work, we propose `\underline{F}actorized \underline{D}iffusion \underline{P}olicies' abbreviated as FDP, a novel policy formulation that enables observational modalities to have differing influence on the action diffusion process by design. This results in learning policies where certain observations modalities can be prioritized over the others such as \texttt{vision>tactile} or \texttt{proprioception>vision}. FDP achieves modality prioritization by factorizing the observational conditioning for diffusion process, resulting in more performant and robust policies. 
Our factored approach shows strong performance improvements in low-data regimes with $15\%$ absolute improvement in success rate on several simulated benchmarks when compared to a standard diffusion policy that jointly conditions on all input modalities. Moreover, our benchmark and real-world experiments show that factored policies are naturally more robust with $40\%$ higher absolute success rate across several visuomotor tasks under distribution shifts such as visual distractors or camera occlusions, where existing diffusion policies fail catastrophically. FDP thus offers a safer and more robust alternative to standard diffusion policies for real-world deployment. Videos are available at~\url{https://fdp-policy.github.io/fdp-policy/}. 
\end{abstract}
\IEEEpeerreviewmaketitle
\section{Introduction}
Humans prioritize different sensory modalities according to the specific requirements of the task \cite{wahn2017attentional}. For instance, Wahn and K\"{o}nig \cite{wahn2017attentional} note that participants engaged in visually demanding tasks are comparatively less receptive to auditory stimuli. They argue that this flexible allocation of human attentional capacity maximizes the capability to process relevant information. Further, humans have been shown to prioritize the more reliable modality between vision and haptics in different situations~\cite{ernst2002humans}. This naturally raises the question of whether policy learning could benefit from such prioritization of observational modalities influencing robot actions. \textbf{Prioritization of the more influential or reliable modality could enable robot policies to learn skills more efficiently and avoid developing spurious correlations with noisy modalities}. Moreover, the number of possible skills is vast and skills may depend more strongly on certain observational modalities over others. For instance, repetitive motions like sweeping are more likely to depend on the robot's proprioception, while locating an object for manipulation is conditioned strongly on its vision. This necessitates a need for an efficient skill learning method that considers the varying levels of influence that different observational modalities may have.

Diffusion models \cite{ho2020denoisingdiffusionprobabilisticmodels} have been extensively leveraged for learning robot skills from demonstrations \cite{chi2023diffusion}. The current methods for training diffusion policies, jointly condition the action diffusion process $\vx$ on all $M$ observational modalities $\vy^{1:M}$ for every task~\cite{chi2023diffusion}. This is a monolithic joint conditioning approach - ``when all you have is a hammer, everything looks like a nail''. Existing diffusion policies do not flexibly accommodate the differing influence of various observational modalities. We empirically show that this joint conditioning approach hurts the sample complexity of diffusion policies as it is difficult to learn the level of dependence that actions have on the observational modalities with limited data. Further, diffusion policies are sensitive to small distribution shifts in \textit{any} modality $\vy^{1:M}$ that they conditions upon, and cannot `de-prioritize' the modality susceptible to noise, similar to humans ~\cite{ernst2002humans}. Incorporating robustness to such shifts require a prohibitively large amount of data when the observation modalities are high-dimensional. 
To that end, we propose a novel policy formulation called Factorized Diffusion Policies (FDP) for enabling prioritization of observational modalities during policy learning. 

At its core, FDP learns a diffusion \textit{base policy} using $k$ ($k < M$) input modalities $\vy^{1:k}$ to be prioritized, followed by a diffusion \textit{residual policy} that learns the noise or score residual conditioned on all modalities $\vy^{1:M}$. We provide a probabilistic formulation to the residual from the first principles, and also develop a novel architecture for efficiently learning it. The base and residual models are then composed to obtain samples from the full conditional action distribution $p(\vx|\vy^{1:M})$. By enabling modality prioritization, FDP introduces flexibility in learning diffusion policies with the inclusion of prioritization order of observational modalities in the hyperparameter search space. Through extensive experiments across visual (\texttt{vision}), point-cloud (\texttt{pcd}), proprioceptive (\texttt{prop}), environment-state (\texttt{env-state}) and tactile (\texttt{tactile}) modalities, we show that leveraging this added flexibility results in more performant and robust diffusion policies. Our contributions are as follows.

\begin{itemize}[leftmargin=*]
    \item We propose FDP, a novel policy formulation that enables observational modalities to have differing influence on the action diffusion process. We mathematically derive a framework to split the jointly conditioned policy into a base policy learned with prioritized modalities and a residual policy learned with all modalities.
    \item We show the merits of modality prioritization through extensive \textbf{experimentation across visual, point-cloud, proprioceptive and tactile} observational inputs on several simulated benchmarks such as RLBench ($15\%\uparrow$), Adroit ($10\%\uparrow$), Robomimic ($10\%\uparrow$) and M3L ($20\%\uparrow$). We thoroughly analyze our method and present ablations.
    \item Our real-world experiments demonstrate the usefulness of FDP in learning policies that are robust to visual distribution shifts ($40\%\uparrow$). Policies learned using the prioritization order of \texttt{prop>vision} were not only \textbf{robust to distractors but also to camera occlusions ($5\times\uparrow$)}, where diffusion policies failed catastrophically.
\end{itemize}

\section{Relevant Work}
\textbf{Sample complexity and Generalization.} Despite recent scaling efforts \cite{kim2024openvla,black2024pi_0,shukor2025smolvla}, the collection of multimodal data is difficult in robotics and the number of variations of tasks is unbounded. Hence several works have tried to improve the sample complexity of learning new skills. Several works leverage compositionality for solving novel combinations of tasks with existing solutions, such as composing learned constraints to generalize to new task combinations in manipulation \cite{liu2024composable} and planning \cite{yang2023compositional}, composing distributions across heterogeneous modalities for tool use \cite{wang2024poco} and sequencing skills for long horizon problems \cite{luo2025generative,mishra2023generative,mishra2024generative}. The most relevant to our work is PoCo \cite{wang2024poco}, that composes single task policies conditioned on different modalities. However, PoCo composes pre-learned policies for the same task and requires manual tuning of the compositional weights. Instead FDP learns the residual to be composed with the base prioritized policy, using the same data and requiring no manual tuning. Augmentation \cite{yu2023scaling,chen2023genaug} or retrieval-based \cite{du2023behavior,memmel2024strap,lin2024flowretrieval} approaches of addressing sample-complexity add a substantial computational and data overhead and are orthogonal to our proposed algorithmic improvement, which may further benefit from them.

\textbf{Residual Learning and Adapters.} Residual reinforcement learning has been used to improve the performance of behavior cloning policies through interaction with the environment \cite{alakuijala2021residual}. These methods learn a residual over the action predicted by the behavior cloned policy through controlled exploration strategies \cite{yuan2024policy}, uncertainty-aware exploration \cite{dodeja2025accelerating}, or as closed-loop corrections for chunked action predictions \cite{ankile2025imitation}, for maximizing the expected returns. Jiang et al. \cite{jiang2024transic} show sim-to-real transfer by learning a supervised residual for human feedback on real world rollouts of policies learned in simulation, maximizing the likelihood of the correction applied. In FDP, the effect of less-influential modalities is captured by learning a residual over a policy trained on prioritized modalities. We theoretically derive this residual within the framework of diffusion and score-based models. Our work is also similar to Q-Adapter \cite{li2025qadaptercustomizingpretrainedllms} in terms of learning a residual using an adapter, but does not necessitate a base foundation model for learning the adapter. 

Several works in robotics have used adapters such as LoRA \cite{hu2022lora} and ControlNet \cite{zhang2023addingconditionalcontroltexttoimage} for fine-tuning multi-task or foundation-models \cite{kim2024openvla,black2024pi_0} on downstream tasks. Prior work has also explored continual adaptation of multi-task policies to novel tasks \cite{liu2023tail,gu2024continual} and adapting pretrained vision \cite{sharma2023lossless} and vision-language models \cite{wen2025tinyvlafastdataefficientvisionlanguageaction} for robotic manipulation. Diff-Control from Liu et al. \cite{liu2024diff} learns a ControlNet with input as the previous action chunk over a diffusion policy base to impart stateful behavior to the policy. Interestingly, FDP can be leveraged to reach a similar learning formulation as Diff-Control, with the residual learned for the previous action chunk instead of a modality. 

\section{Background}
\label{sec:bkg}
\textbf{Diffusion Models.} Gaussian diffusion models \cite{sohldickstein2015deep} learn the reverse diffusion kernel $p_\vtheta(\vx_{t-1}|\vx_{t})$ for a fixed forward kernel that adds Gaussian noise at each step $p(\vx_t|\vx_{t-1})= \mathcal{N}(\vx_t; \sqrt{\alpha_t}\vx_{t-1}, (1-\alpha_t)\mathcal{I})$, such that $p(\vx_T)\approx\mathcal{N}(0, \mathcal{I})$. Here, $t<=T$ is the diffusion time step and $\alpha_t$ is the noise schedule. In practice, the models learn a reparametrized form corresponding to the noise added to the input $\vepsilon_\vtheta(\vx_t, t)$\cite{ho2020denoisingdiffusionprobabilisticmodels}.

\textbf{Score-based Models.} Song et al. \cite{song2020score} presented a unified framework showing that both diffusion models~\cite{sohldickstein2015deep,ho2020denoisingdiffusionprobabilisticmodels} and score-based models~\cite{song2020generative} can be interpreted as discretizations of different forward stochastic differential equations (SDEs). The latter learn the score $\nabla_{\vx_t} \log p_{\sigma_t}(\vx_t)$ at different noise scales $\sigma_t$ required for sampling from the data distribution. Explicit Score Matching (ESM) \cite{hyvarinen2005estimation, vincent2011connection} was proposed to estimate the score by minimizing the Fisher divergence with the Gaussian-smoothed data distribution $p_{\sigma_t}(\vx_t){=}\int p(\vx)\mathcal{N}(\vx_t;\vx,\sigma_t^2I)d\vx$. Denoising Score Matching (DSM) alleviates the computational difficulties of ESM \cite{vincent2011connection,chao2022denoisinglikelihoodscorematching}, and is shown in Equation \ref{eq:score}, where $s_\vtheta(\vx_t,\sigma_t)$, abbreviated to $s_\vtheta(\vx_t)$, represents the learned score model.
\begingroup
\allowdisplaybreaks
\begin{equation}
    \label{eq:score}
    \scalebox{1}{
    \parbox{\linewidth}{$
    \begin{aligned}
        &\mathcal{J}_{\sigma_t}(\vtheta) 
        \stackrel{\text{ESM}}{=}
        \mathbb{E}_{p_{\sigma_t}(\vx_t)}\left[
        \frac{1}{2}\left\|\nabla_{\vx_t} \log p_{\sigma_t}(\vx_t) - s_\vtheta(\vx_t)\right\|^2_2
        \right] \\
        &\stackrel{\text{DSM}}{=}
        \mathbb{E}_{p_{\sigma_t}(\vx, \vx_t)}\left[
        \frac{1}{2}\left\|\nabla_{\vx_t} \log p_{\sigma_t}(\vx_t|\vx) - s_\vtheta(\vx_t)\right\|^2_2
        \right] + C 
    \end{aligned}
    $}}
\end{equation}
\endgroup
Diffusion models use a $(1{-}\Bar{\alpha_t})$ weighted DSM objective along with a forward transition kernel $p_{\bar{\alpha}_t}(\vx_t|\vx){=}\mathcal{N}(\vx_t;\sqrt{\Bar{\alpha_t}}\vx,(1{-}\Bar{\alpha_t})I)$ with discrete time and $\bar{\alpha_i}{=}\prod_{j=1}^i\alpha_j$, yielding the simplified diffusion loss from Ho et al. \cite{ho2020denoisingdiffusionprobabilisticmodels}. Score-based model typically use $\mathcal{N}(\vx_t;\vx,\sigma_t^2I)$, where $\alpha_t$ and $\sigma_t$ are respective noise scales. In the simplified case, an optimal diffusion model is related to the score of the $\alpha_t$-diffused data distribution by $-\vepsilon^*_\vtheta(\vx_t, t)/\sqrt{1-\Bar{\alpha_t}} \stackrel{def}= \vs^*_\vtheta(\vx_t) {=} \nabla_{{\vx}_t} \log p(\vx_t)$ \cite{song2020score}. Typically, diffusion models generate samples via progressive denoising through the reverse diffusion process \cite{ho2020denoisingdiffusionprobabilisticmodels}, while score-matching models sample from the data distribution using Langevin dynamics \cite{roberts1996exponential}.

\textbf{Classifier Guided Sampling.} Dhariwal and Nichol \cite{dhariwal2021diffusion} obtain conditional samples from an unconditional diffusion model trained on $\vx$ using Bayes' theorem. We can sample from a class $\vy$ by decomposing the conditional score at time step $t$ into the unconditional score and the classifier gradient $\nabla_{\vx_t} \log p(\vx_t|\vy) = \nabla_{x_t} \log p(\vx_t;\vtheta) + \nabla_{x_t} \log p(\vy|\vx_t;\vphi)$.
However, classifier guidance needs an explicit classifier trained on noisy samples to estimate the gradients \cite{ho2022classifier}.

\begin{figure}[t]
    \centering
    \includegraphics[width=\linewidth]{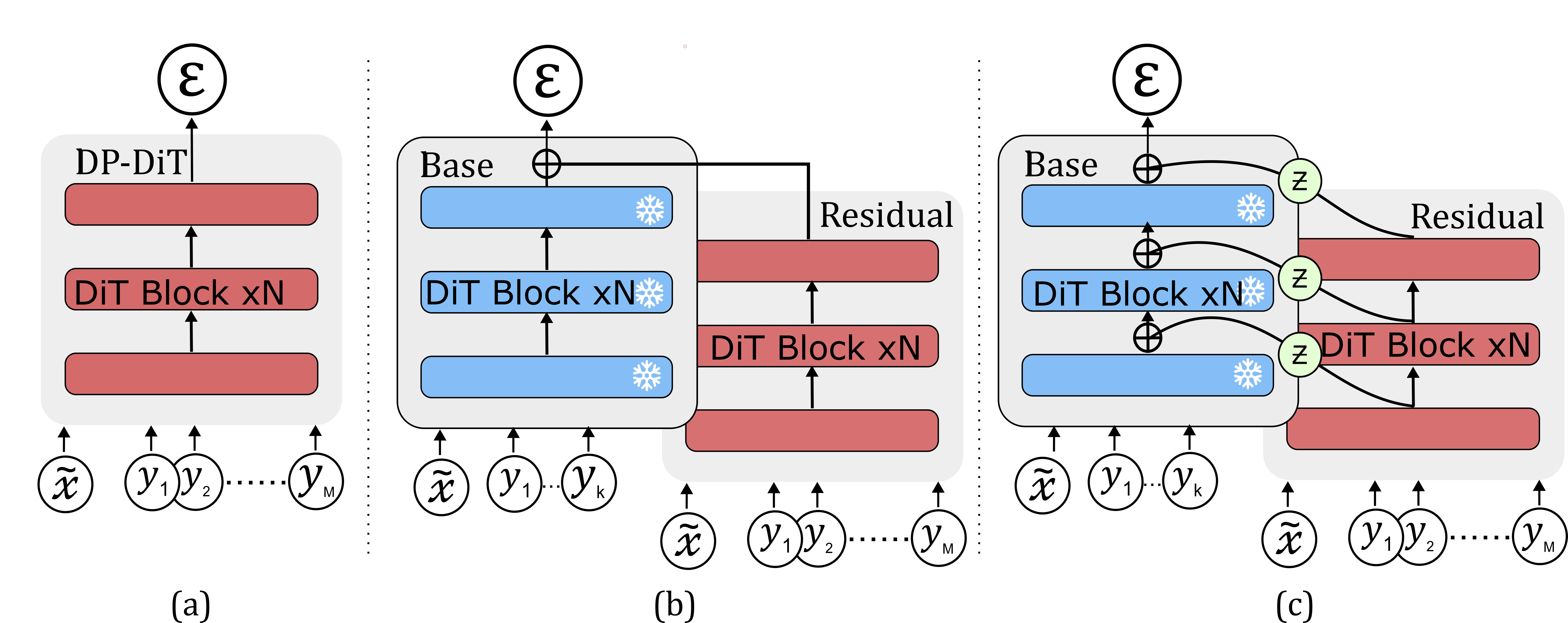}
    \caption{Architectural representations for [a] diffusion policy that jointly conditions on all observational modalities, [b] straightforward composition of the score outputs from $\pi_\text{base}$ and $\pi_\text{res}$ and [c] FDP architecture with block-wise composition with a layer $\mathcal{Z}$ applied on $\pi_\text{res}$.}
    \label{fig:archie}
\vspace{-0.5cm}
\end{figure}

\section{Methodology}
Assume that we have robot demonstrations $D=\{(\vx, \vy)_i\}$ where $i=1..N$, consisting of actions $\vx$ and different observational modalities $\vy^{1:M}$, such as images, point clouds, proprioception or tactile. Let $\vy^{1:k}$ be the prioritized observational modalities of the M total modalities, where $\vy^{1:k} {\equiv} \vy^{1},..,\vy^{k}$ and $k {<} M$. 
To sample actions $\vx$ conditioned on all $\vy^{1:M}$, we need to estimate the score $\nabla_{\vx_t}\log p(\vx_t|\vy^{1:M})$. Taking a cue from classifier guidance \cite{dhariwal2021diffusion}, we decouple the observational modalities utilizing Bayes' theorem to obtain Equation \ref{eq:bayes}. For scores to be valid, observational modalities $\vy^{1:M}$ can be noised with a Gaussian kernel $\mathcal{N}(\tilde{\vy};\vy,\tau^2I)$ of variance $\tau^2$ that is small enough such that $p_{\tau}(\tilde{\vy}) \approx p(\vy)$, where we drop the notation $\tau$ going forward. Actions $\vx$ are noised with the kernel $p_{\bar{\alpha}_t}(\vx_t|\vx)$.
\begingroup
\allowdisplaybreaks
\begin{equation}
    \label{eq:bayes}
    \scalebox{1}{
    \parbox{\linewidth}{$
    \begin{aligned}
    &\vs^*(\vx_t,\vy^{1:M}) = \nabla_{\vx_t}\log p(\vx_t|\vy^{1:M}) \\
    &= \nabla_{\vx_t} \log p(\vx_t|\vy^{1:k})
    + \nabla_{\vx_t} \log p(\vy^{k+1:M}|\vx_t,\vy^{1:k})
    \end{aligned}
    $}}
\end{equation}
\endgroup
The first score term $\nabla_{\vx_t} \log p(\vx_t|\vy^{1:k})$ corresponds to a policy $\vs_\vtheta(\vx_t,\vy^{1:k})$ that would be obtained on training with just the modalities $\vy^{1:k}$, referred to as $\pi_\text{base}$ going forward.
The effect of the other modalities is captured in the second score term $\nabla_{\vx_t} \log p(\vy^{k+1:M}|\vx_t,\vy^{1:k})$. However, explicitly training a classifier $p(\vy^{k+1:M}|\vx_t,\vy^{1:k})$ as suggested by Dhariwal and Nichol \cite{dhariwal2021diffusion} is impractical due to the high dimensionality and continuity of multiple observational modalities $\vy^{1:M}$, such as images and tactile. Hence, we employ explicit score matching $\mathcal{J}_{\alpha_t}(\vphi)$ \cite{hyvarinen2005estimation,vincent2011connection} as shown in Equation \ref{eq:esm}.
\begingroup
\allowdisplaybreaks
\begin{equation}
\label{eq:esm}
\scalebox{1}{
\parbox{\linewidth}{$
\begin{aligned}
\mathbb{E}_{p_{\alpha_t}(\vx_t, \vy^{1:M})} \Biggl[
\frac{1}{2}
\left\|
\begin{array}{l}
 \nabla_{\vx_t} \log p_{\alpha_t}(\vy^{k+1:M}|\vx_t,\vy^{1:k}) \\
 - \vs_\vphi(\vx_t, \vy^{1:M})
\end{array}
\right\|^2_2
\Biggr]
\end{aligned}
$}}
\end{equation}
\endgroup
Due to the high computational complexity of estimating the empirical scores such as $\nabla_{\vx_t} \log p_{\alpha_t}(\vy^{k+1:M}|\vx_t,\vy^{1:k})$, Chao et al. \cite{chao2022denoisinglikelihoodscorematching} derive the denoising likelihood score matching (DLSM) objective for conditional distributions, which forms the basis for our next result.
\begin{theorem}
    \label{thm:1}
    Explicit score matching for $\vs_\vphi(\vx_t, \vy^{1:M})$ in Equation \ref{eq:esm} is equivalent to the objective $\mathcal{J}^{res}_{\alpha_t}(\vphi)$:
    \begin{equation}
    \label{eq:codig-loss}
    \scalebox{1}{
    \parbox{\linewidth}{
    \begin{align*}
        \mathbb{E}_{\stackrel{p_{\alpha_t}(\vx, \vx_t,}{\vy^{1:M})}}
        \Biggl[
        \frac{1}{2} \left\| \nabla_{\vx_t} \log p_{\alpha_t}(\vx_t|\vx) {-}
        \begin{array}{l}
        \vs^*(\vx_t,\vy^{1:k}) \\
        - \vs_{\vphi}(\vx_t, \vy^{1:M})
        \end{array}
        \right\|^2_2 \Biggr]
    \end{align*}
    }}
    \end{equation}
\end{theorem}
Here $\vs^*(\vx_t,\vy^{1:k})$ is the the frozen optimal score model for $\nabla_{\vx_t} \log p(\vx_t|\vy^{1:k})$, approximated using a learned model $\pi_\text{base}:\vs_{\vtheta}(\vx_t,\vy^{1:k})$ in practice. \\

We believe FDP to be the first work to prove this equivalence for an arbitrary number of conditionals and directly learn the classifier guidance in a high-dimensional setting for the purposes of policy learning. Interestingly, comparison of $\mathcal{J}^{res}_{\alpha_t}(\vphi)$ with DSM shown in Equation \ref{eq:score} reveals that the effect of $\vy^{k+1:M}$ captured through $\vs_{\vphi}(\vx_t, \vy^{1:M})$ can be learned as a residual to $\pi_\text{base}:\vs_{\vtheta}(\vx_t,\vy^{1:k})$, the policy learned with just the modalities $\vy^{1:k}$. Thus we refer to $\vs_{\vphi}(\vx_t, \vy^{1:M})$ as $\pi_\text{res}$. \textbf{Essentially, by factorizing the score of the full-conditional action, FDP learns $\pi_\text{base}$ using $\vy^{1:k}$, and then learns $\pi_\text{res}$ using $\vy^{1:M}$ and a frozen $\pi_\text{base}$}. This two-phase training prioritizes modalities $\vy^{1:k}$ over $\vy^{k+1:M}$.  \\

A concise proof of Theorem \ref{thm:1} is presented, and a detailed version can be found on our website. We substitute $\vs_{\vphi}(\vx_t, \vy^{1:M})$ as $\scalemath{1}{\bullet}$ for brevity. The inner product obtained by opening the square in Equation \ref{eq:esm} can be simplified as- 
\begingroup
\allowdisplaybreaks
\begin{align}
    &\scalemath{1}{\mathbb{E}_{p_{\alpha}(\vx_t, \vy^{1:M})}\left[
    \left\langle\bullet,\nabla_{\vx_t} \log p_{\alpha_t}(\vy^{k+1:M}
    |\vx_t,\vy^{1:k}) \right\rangle \right]} = \label{eq:proof-res1}\\
    &\scalemath{1}{\mathbb{E}_{\stackrel{p_{\alpha}(\vx_t,}{\vy^{1:M})}}\left[
    \left\langle\bullet,\nabla_{\vx_t} \log p_{\alpha_t}(\vx_t|\vy^{1:M}) {-} \nabla_{\vx_t} \log p_{\alpha_t}(\vx_t|\vy^{1:k}) \right\rangle \right]} \notag
\end{align}
\endgroup
Further simplifying the inner product with the first term- 
\begingroup
\allowdisplaybreaks
\begin{align}
    &\scalemath{1}{\mathbb{E}_{p_{\alpha_t}(\vx_t, \vy^{1:M})}\big[\langle\vs_{\vphi}(\vx_t, \vy^{1:M}),\nabla_{\vx_t} \log p_{\alpha_t}(\vx_t|\vy^{1:M})\rangle\big]}  \notag \\
    &{=}\scalemath{1}{\mathbb{E}_{p(\vy^{1:M})}\!\!\int_{\vx_{t}}\!\!\!\!p_{\alpha}(\vx_t|\vy^{1:M}) \langle\bullet, 
    \frac{\nabla_{\vx_t} p_{\alpha_t}(\vx_t|\vy^{1:M})}{p_{\alpha_t}(\vx_t|\vy^{1:M})}\rangle d\vx_t} \notag \\
    &{=}\scalemath{1}{\mathbb{E}_{p(\vy^{1:M})}\!\!\int_{\vx_{t}} \!\!\!\!
    \langle\bullet,\!\! \nabla_{\vx_t} \!\!\! \int_{\vx_0} \!\!\!\! p_{0}(\vx_0|\vy^{1:M}) p_{\alpha_t}(\vx_t|\vx_0,\vy^{1:M})d\vx_0 \rangle d\vx_t} \notag \\
    &{=}\scalemath{1}{\mathbb{E}_{p_{\alpha_t}(\vx_0, \vx_t, \vy^{1:M})}\big[\langle\bullet, \nabla_{\vx_t} \log p_{\alpha_t}(\vx_t|\vx_0) \rangle\big]} \label{eq:proof-res2}
\end{align}
\endgroup

Note that $\scalemath{1}{\mathbb{E}_{\stackrel{p_{\alpha_t}(\vx_t,}{\vy^{1:M})}}\left[\frac{1}{2}||\nabla_{\vx_t} \log p_{\alpha_t}(\vy^{k+1:M}|\vx_t,\vy^{1:k})||^2_2\right]}$ is a constant. Substituting results obtained in Equations \ref{eq:proof-res1} and \ref{eq:proof-res2} back in Equation \ref{eq:esm}, and adding the constant term $\scalemath{1}{\mathbb{E}_{\stackrel{p_{\alpha_t}(\vx_t,}{\vy^{1:k})}}\left[\frac{1}{2}||\nabla_{\vx_t} \log p_{\alpha_t}(\vx_t|\vy^{1:k}) - \nabla_{\vx_t} \log p_{\alpha}(\vx_t|\vx_0)||^2_2\right]}$ we complete the proof for Theorem \ref{thm:1}. Hence, we prove that the ESM $\mathcal{J}_{\alpha_t}(\vphi)$ in Equation \ref{eq:esm} is equivalent to minimizing the objective $\mathcal{J}^{res}_{\alpha_t}(\vphi)$ presented in Theorem \ref{thm:1}, differing up to a constant. \\ 

Theorem \ref{thm:1} implies that the effect of de-prioritized modalities $\vy^{k+1:M}$ can be learned as a residual over the prioritized modalities $\vy^{1:k}$. From a different lens, $\pi_\text{res}$ effectively learns the classifier guidance required to sample from $\pi_\text{base}$ further conditioned on $\vy^{k+1:M}$. \textbf{Learning $\pi_\text{res}$ as a residual of $\pi_\text{base}$ ensures that the policy does not overfit modalities $\vy^{k+1:M}$, but only learns correlations to bridge the error arising from $\pi_\text{base}$ trained on the prioritized modalities $\vy^{1:k}$}. Hence, policies learned in this factorized way are naturally robust to distribution shifts in $\vy^{k+1:M}$. Moreover, explicit prioritization of $y^{1:k}$ by training $\pi_\text{base}$ prior to learning the residual leads to sample efficiency, as the model learns correlations with the stronger modality without having to attend to other modalities. 

\noindent\textbf{Factorizing Diffusion Policies}. As developed in Section \ref{sec:bkg}, Theorem \ref{thm:1} applies to diffusion models. Resolving $\nabla_{\vx_t} \log p_{\alpha_t}(\vx_t|\vx)$ to $-\vepsilon_0/\sqrt{1-\Bar{\alpha_t}}$ and replacing $\vs^*_\vtheta(\vx_t)$ with $-\vepsilon_{\theta}(\vx_t, t)/\sqrt{1-\Bar{\alpha_t}}$, we get the simplified diffusion losses for $\pi_\text{base}$ and $\pi_\text{res}$.
\begingroup
\allowdisplaybreaks
\begin{equation}  
\scalebox{1}{$
\begin{aligned}
    \mathcal{L}^t_{base}(\vtheta) 
    &= \mathbb{E}_{p(\vx_0,\vy^{1:k})\mathcal{N}(\vepsilon_0;0,\mathcal{I})}
    \left[\left\|\vepsilon_0 - \vepsilon_\vtheta(\vx_t,\vy^{1:k},t)\right\|_2^2\right] \label{eq:mm-loss} \\
    \mathcal{L}^t_{res}(\vphi) 
    &= \mathbb{E}_{\stackrel{\vx_0, \vy^{1:M} \sim p}{\vepsilon_0 \sim \mathcal{N}(0, \mathcal{I})
    }} \Biggl[
    \frac{1}{2} \left\| 
    \begin{array}{l}
    \vepsilon_0 - \vepsilon_{\vtheta}(\vx_t, \vy^{1:k}, t) \\
    - \vepsilon_{\vphi}(\vx_t, \vy^{1:M}, t)
    \end{array}
    \right\|^2_2 \Biggr]
\end{aligned}$}
\end{equation}
\endgroup

From the perspective of diffusion models, $\pi_\text{base}$ maximizes a reweighted lower bound on the data likelihood only considering the prioritized $k$ modalities, while $\pi_\text{res}$ learns a residual over $\pi_\text{base}$ to maximize it for demonstration data with all the modalities included, thus learning their residual effect. Since diffusion models are trained on discrete time steps, $\pi_\text{res}$ is learned on the same time discretization as used for $\pi_\text{base}$. Once trained, actions can be sampled from the conditional distribution $p(\vx|\vy^{1:M})$ using reverse diffusion \cite{ho2020denoisingdiffusionprobabilisticmodels} on the composition \cite{du2023reduce} of $\pi_\text{base}$ and $\pi_\text{res}$:
\begingroup
\allowdisplaybreaks
\begin{equation}\label{eq:diff-sampling-codig}
\scalebox{1}{$
\begin{aligned}
&\vx_{t-1} {\sim} \mathcal{N}\!\left(\vx_t;\tfrac{1}{\sqrt{\alpha_t}}\left(\vx_t {-} \tfrac{1-\alpha_t}{\sqrt{1-\bar{\alpha}_t}}\vepsilon(\vx_t,\vy^{1:M},t)\right), \sqrt{1{-}\alpha_t}\,\mathcal{I}\right) \\[0.5ex]
&\vepsilon(\vx_t,\vy^{1:M},t) = \vepsilon_\vtheta(\vx_t,\vy^{1:k},t) + \vepsilon_\vphi(\vx_t,\vy^{1:M},t)
\end{aligned}
$}
\end{equation}
\endgroup
\textbf{Architectural Implementation of FDP.}
The models $\pi_{\text{base}}$ and $\pi_{\text{res}}$ can be instantiated using standard architectures such as UNet \cite{ronneberger2015u} or DiT \cite{peebles2023scalablediffusionmodelstransformers}.
 Inspired from the late stage score-composition, we propose a more integrated way to compose $\pi_\text{base}$ and $\pi_\text{res}$, as shown in Figure \ref{fig:archie}. Instead of learning a residual for the final score output, $\pi_\text{res}$ learns the blockwise residual over the intermediate outputs of the frozen $\pi_\text{base}$. Specifically, let $\mathcal{F}^i_{\text{base}}$ and $\mathcal{F}^i_{\text{res}}$ denote the $i$-th DiT block outputs of the base and residual models, respectively. Then the composed output at level $i$ can be written as $\scalemath{1}{\mathcal{F}^i_{base}(\vx', \vy'^{1:k}) + \mathcal{Z}(\mathcal{F}^i_{res}(\vx', \vy'^{1:M}))}$, where $x'$ and $y'^{1:M}$ are layer inputs. Similar to Zhang et al. \cite{zhang2023addingconditionalcontroltexttoimage}, $\mathcal{Z}$ is a zero-initialized layer to avoid harmful updates at the start of the training and to ensure that gradient updates to the residual model improve the predictions of the composed model over $\pi_\text{base}$. 
 This architecture enables a simplified training objective from Ho et al. \cite{ho2020denoisingdiffusionprobabilisticmodels} for the residual model. Our residual model is structured following the Vision Transformers architecture \cite{dosovitskiy2020image}. In $\pi_{\text{res}}$, all observational modalities are passed through self-attention layers after encoding into a single embedding. 

All transformer-based models are trained over 2000 epochs with a batch size of 64 for visual tasks and for 3000 epochs with a batch size of 256 for low-dimensional tasks. All models except VLAs are trained on NVIDIA A5000 GPUs, with training times ranging from 6-12 hours depending on model size and the number of camera inputs. Model \texttt{prop} $\pi_{base}$ consists of ${\sim}30M$ parameters, while \texttt{vis} $\pi_{res}$ with two camera image inputs is ${\sim}55M$ parameters large. Our current implementations support an action prediction latency of $\sim$50ms for transformer-based diffusion policy baseline, $\sim$100ms for UNet \cite{chi2023diffusion} and output composition of models as shown in Figure \ref{fig:archie} $[b]$, and $\sim$150ms for FDP model in $[c]$.

\section{Simulation Experiments}
\label{sec:sim-res}
We train and evaluate FDP and related baselines on 14 tasks from RLBench \cite{rlbench} along with their distractor variants, 4 tasks from Adroit \cite{rajeswaran2017learning}, 4 tasks from Robomimic \cite{mandlekar2021matters}, and the visuo-tactile insertion task from M3L \cite{sferrazza2024power}. RLBench provides a diverse suite of visuomotor manipulation tasks  with joint positions as the action space and a built-in planner for demonstration collection. Within RLBench, we evaluate prioritization orders \texttt{prop}$\lessgtr$\texttt{vision} and \texttt{pcd}$\lessgtr$\texttt{vision} for FDP. The Adroit benchmark contains high-dimensional hand manipulation tasks performed with a 24-DoF anthropomorphic hand. Environments in Robomimic use an action representation defined by changes in end-effector position and orientation (axis–angle). For both Adroit and Robomimic, we explore the prioritization orders \texttt{prop}$\lessgtr$\texttt{env-state}. Finally, the M3L environment requires precise insertion of differently shaped pegs into holes randomly placed on a surface, using a single RGB camera and two tactile sensors for perception and $\Delta$xyz as the action representation. Demonstrations for M3L are collected using an expert RL policy proposed by Sferrazza et al. \cite{sferrazza2024power}. On M3L we evaluate the performance of \texttt{vision}$\lessgtr$\texttt{tactile} prioritizations using FDP against jointly conditioned diffusion policy. Unless otherwise noted, reported results are averaged over 300 rollouts. Additional experimental details are provided on our project webpage. ~\url{https://fdp-policy.github.io/fdp-policy/}.

\begin{figure}[t]
    \centering\includegraphics[width=\linewidth]{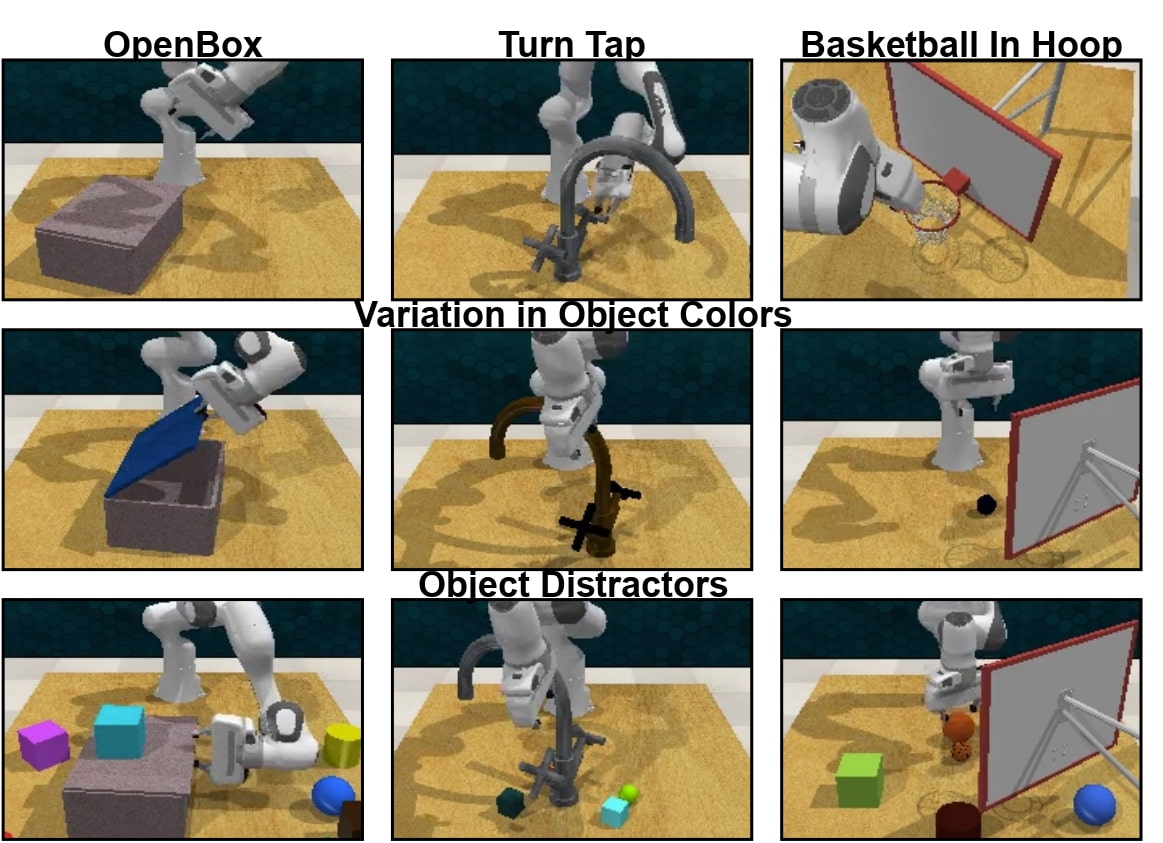}
    \caption{We modify the original RLBench environments to introduce color variations and add distractors.}
    \label{fig:rlbench_dist}
\end{figure}

\begin{figure}
    \centering
    \includegraphics[width=1\linewidth]{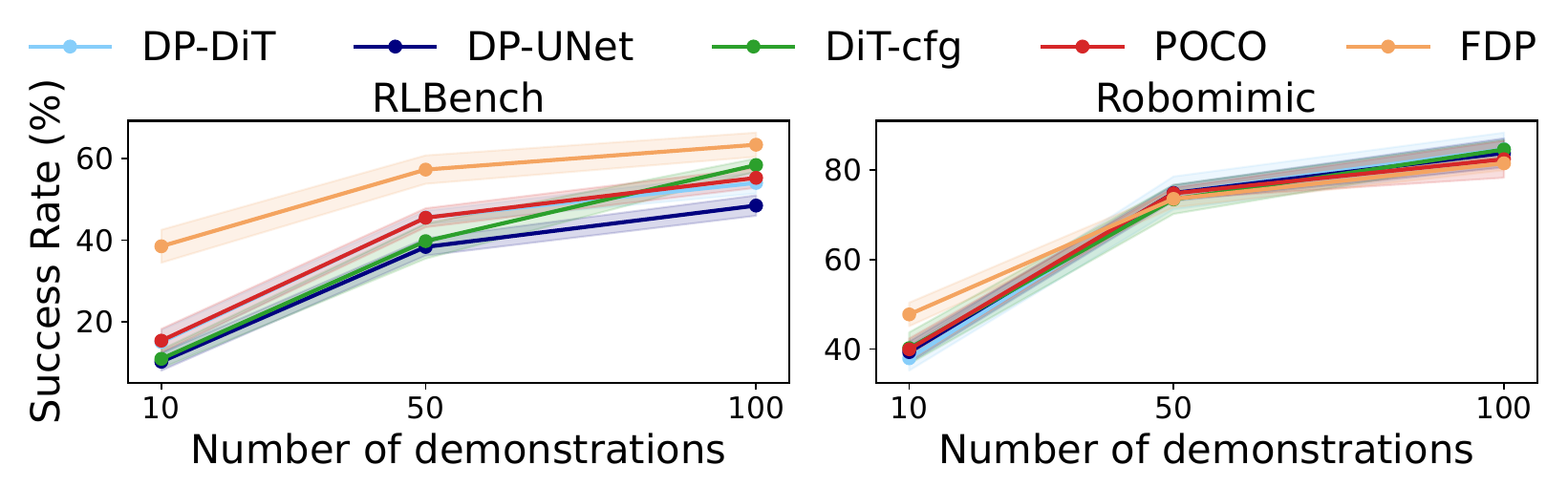}
    \caption{Mean (with Std. Dev.) Performance for all models across 10-50-100 demonstrations on RLBench and Robomimic. Prioritization of modalities using FDP enables strong performance gains, especially in low-data regimes.}
    \label{fig:res1-scaling}
\end{figure}
\begin{figure}
    \centering
    \includegraphics[width=1\linewidth]{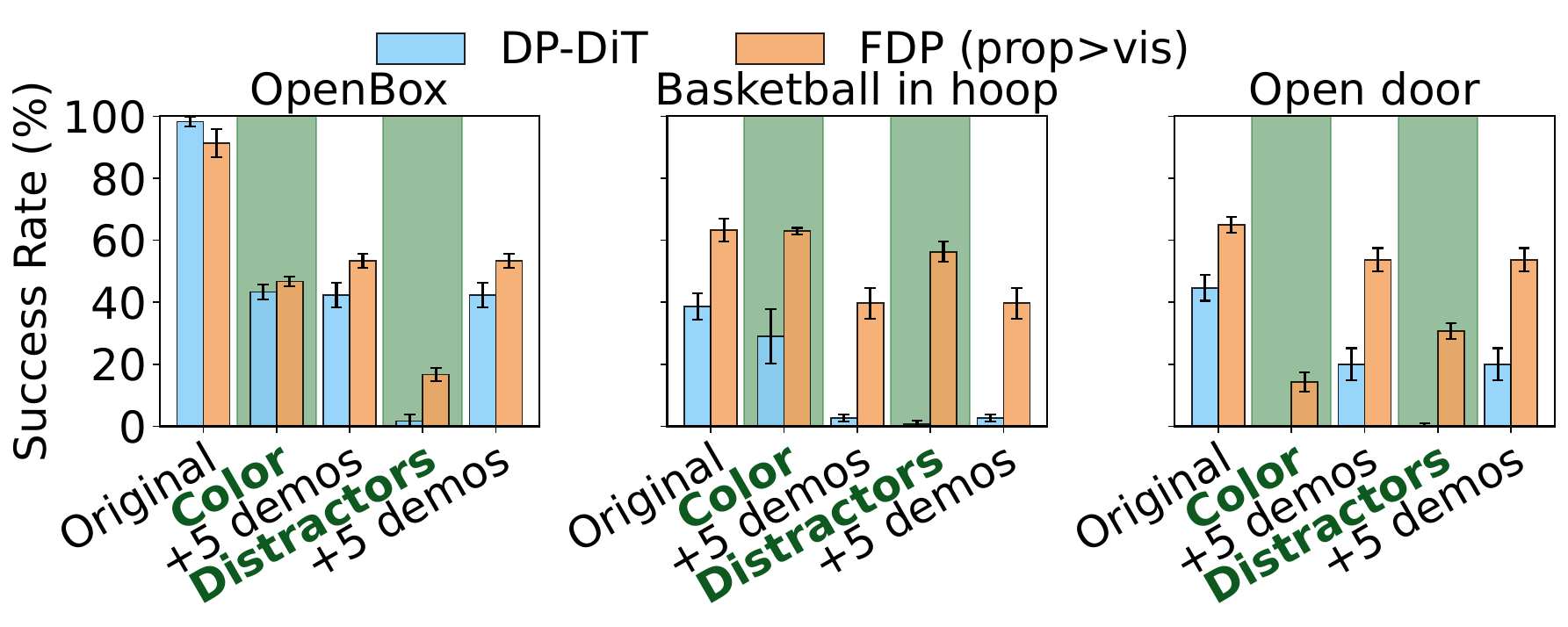}
    \caption{Performance of FDP (\texttt{prop>vision}) and DP-DiT across original, color-variant and distractor environments. We also fine-tune on 5 additional demos collected in the modified environments. Note the strong performance of FDP (\texttt{prop}>\texttt{vision}) in the color-variant and distractor experiments.}
    \label{fig:placeholder}
\end{figure}

\begin{figure*}[t]
    \centering
    \subfloat[\label{fig:a}RLBench tasks with 10 demonstrations]{%
        \includegraphics[width=0.30\textwidth]{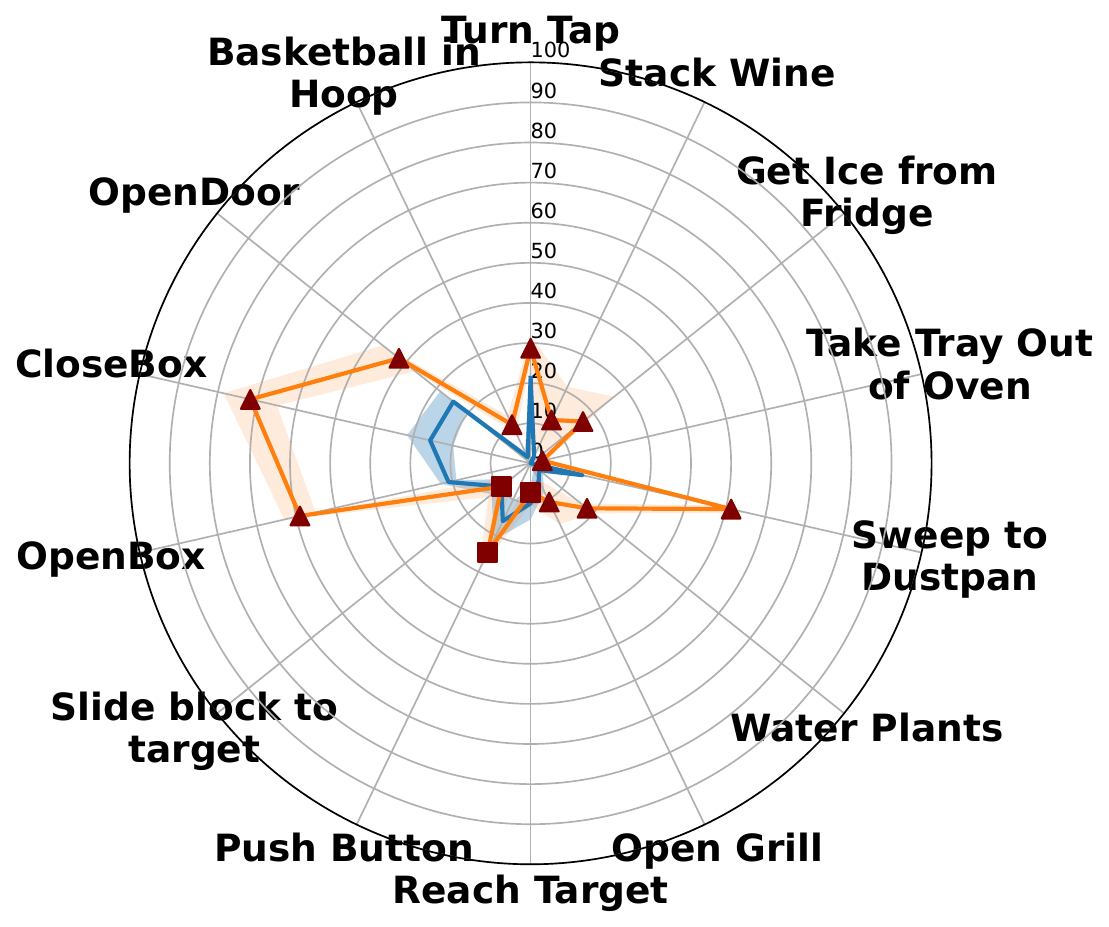}}
    \hfill
    \subfloat[\label{fig:b}RLBench tasks with 50 demonstrations]{%
        \includegraphics[width=0.30\textwidth]{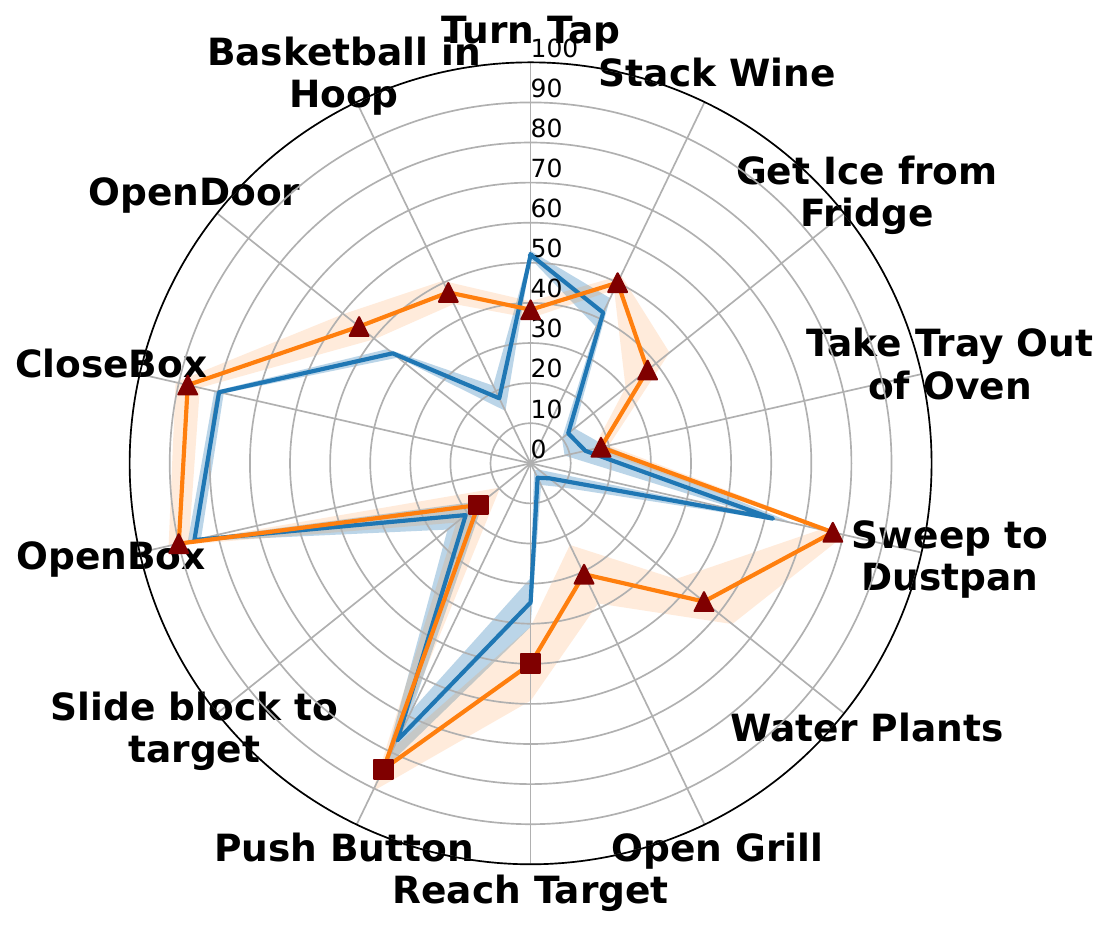}}
        \hfill
    \subfloat[\label{fig:c}Robomimic \& Adroit with 10-50 demos]{%
        \includegraphics[width=0.30\textwidth]{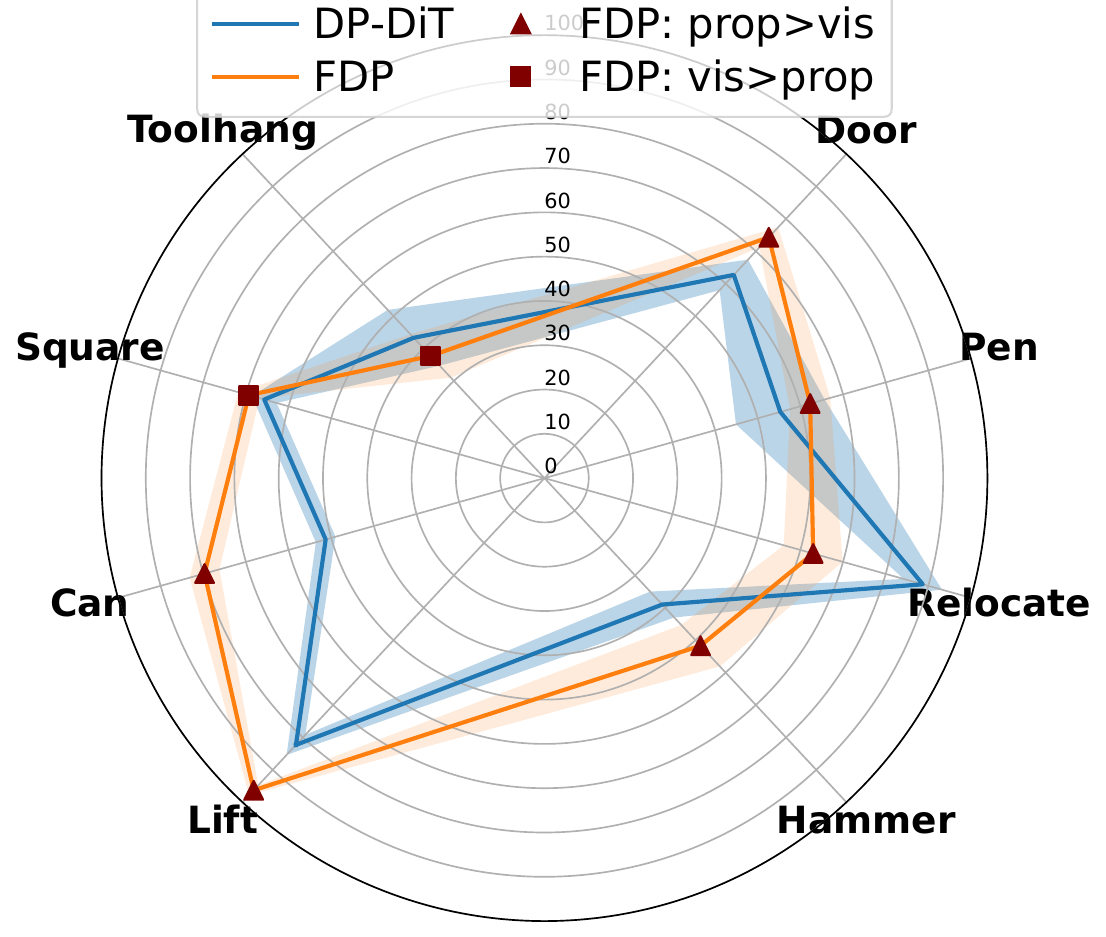}}
    \caption{Radial plot showing performance of FDP in comparison to DP-DiT. For FDP, the best results obtained using $\blacksquare$ \texttt{vis>prop} and $\blacktriangle$ \texttt{prop>vision} are marked on the plots. These plots show that searching through the modality prioritization space yields improvements in a wide-spectrum of tasks.}
    \label{fig:rlbench_adroit}
\end{figure*}

\noindent\textbf{Baselines.} For evaluation of sample efficiency in visuomotor tasks, we compare against several approaches that differ in the way in which they model generative policy learning.
However, for all approaches, we choose DiT-small ($\sim$90M) \cite{peebles2023scalablediffusionmodelstransformers} as our model architecture. We implement Diffusion Policy \cite{chi2023diffusion} using DiT, referred to as DP-DiT in our results. For comparison, we also include UNet \cite{ronneberger2015u} implemented by Chi et al. \cite{chi2023diffusion} in our baselines as DP-UNet. We reformulate POCO \cite{wang2024poco} to compose observational modalities.  We train $\pi_\text{base}$ and $\pi_\text{res}$ models independently, prior to sampling from the composed distribution \cite{du2023reduce} using $\scalemath{1}{\vepsilon(\vx_t,\vy^{1:M},t)} {=} \scalemath{1}{\vepsilon_\vtheta(\vx_t,\vy^{1:k},t) + \lambda*\vepsilon_\vphi(\vx_t,\vy^{1:M},t)}$. Here, $\lambda=0.1$ based on POCO's ablations~\cite{wang2024poco}. We also report results for classifier-free guidance \cite{ho2022classifierfreediffusionguidance} as CFG, where we train a single model and switch out the weaker modality with a probability of 0.2. We then sample using $\scalemath{1}{\vepsilon(\vx_t,\vy^{1:M},t)} = \scalemath{1}{\lambda_1*\vepsilon_\vtheta(\vx_t,\vy^{1:k}, \vy^{k+1:M},t) + \lambda_2*\vepsilon_\vtheta(\vx_t,\vy^{1:k}, \vphi,t)}$, where we set $\lambda_1=1.1$ and $\lambda_2=0.1$, as suggested by \cite{ho2022classifierfreediffusionguidance}. We also fine-tune a 450M parameter vision-language action model SmolVLA \cite{shukor2025smolvla} for at least 40k steps, evaluating its sample-efficiency and robustness on selected RLBench tasks. For real-world and distractor experiments in simulation, we compare against DP-DiT.

\begin{figure}
    \centering
    \includegraphics[width=1\linewidth]{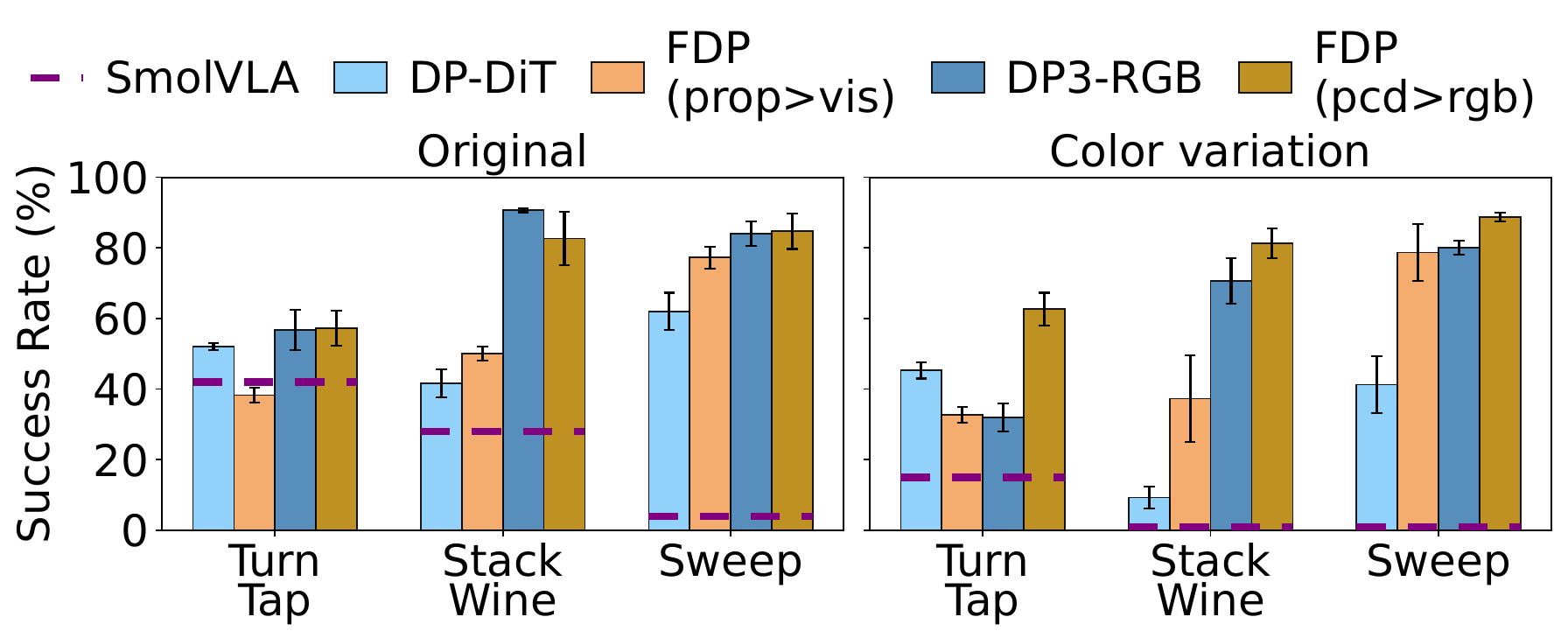}
    \caption{Evaluation of robustness gained using \texttt{prop|pcd>vision}. Notably, we see a significant drop in performance of DP-DiT and DP3 on introducing color variations in the task. Unlike FDP, SmoLVLA despite having a strong VLM backbone, fails to adapt to color variations across tasks.}
    \label{fig:vla-dp3}
\end{figure}

\subsection{Sample Efficiency Gain Through Modality Prioritization.}
\begin{figure}[t]
\centering

\begin{minipage}[t]{0.56\columnwidth}
\vspace{0pt} 
\scriptsize
\setlength{\tabcolsep}{2pt}
\renewcommand{\arraystretch}{0.92}
\begin{tabular}{@{}l c c c c@{}}
\hline
\textbf{Task} & \textbf{Demos} &
\shortstack{\textbf{DP-DiT}} &
\shortstack{\textbf{vision>}\\\textbf{tactile}} &
\shortstack{\textbf{tactile>}\\\textbf{vision}} \\
\hline
\multirow{2}{*}{square peg}   & 100 & 22 & \textbf{48} & 28 \\
                              & 200 & 52 & \textbf{72} & 24  \\ \hline
\multirow{2}{*}{triangle peg} & 100 & 14 & \textbf{42} & 10  \\
                              & 200 & 28 & \textbf{50} & 16  \\ \hline
All pegs                      & 200 &  6 & \textbf{26} &  4  \\
                              
\hline
\end{tabular}
\end{minipage}\hfill
\begin{minipage}[t]{0.42\columnwidth}
\vspace{2pt} 
\scriptsize
\setlength{\tabcolsep}{1pt}
\renewcommand{\arraystretch}{0.92}
\begin{tabular}{@{}cc@{}}
\includegraphics[width=0.49\linewidth]{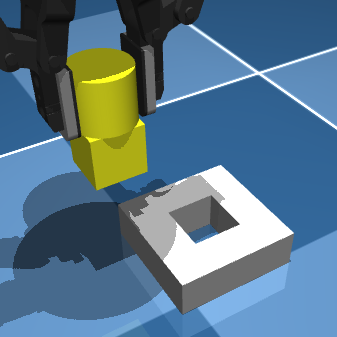} &
\includegraphics[width=0.49\linewidth]{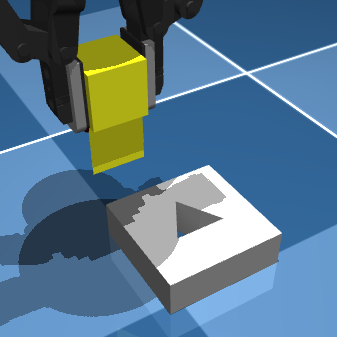} \\
\end{tabular}
\end{minipage}
\caption{Results from the visuo-tactile insertion tasks from M3L \cite{sferrazza2024power}. We see a clear benefit with the prioritization order of \texttt{vision}>\texttt{tactile}. The vision modality plays a crucial role in navigating the peg towards the hole~\cite{zhao2025touchbeginsvisionends}. Tactile input becomes useful once the peg is in contact with the contour of the hole, and is learned as a residual over vision $\pi_\text{base}$.}
\label{tab:m3l_side}
\end{figure}

FDP enables us to include the prioritization order of modalities for tasks as a hyperparameter. Prioritizing either \texttt{vision}, \texttt{env-state}, or \texttt{prop} modalities using \textbf{FDP consistently outperforms joint conditioning across a wide range of visuomotor, state-based and tactile tasks} in RLBench, Robomimic, Adroit (Fig. \ref{fig:rlbench_adroit}) and M3L (Table \ref{tab:m3l_side}) respectively. In RLBench, modality prioritization with FDP achieves on average a $15\%$ higher success rate with $10$ and $50$ demonstrations, and a $10\%$ higher success rate with $100$ demonstrations, compared to the strongest baseline. We also observe clear gains with the prioritization order of \texttt{prop>env-state} in Robomimic and Adroit tasks. Fig. \ref{fig:res1-scaling} illustrates how performance scales with increasing demonstrations in RLBench and Robomimic. Prioritization is \textbf{especially beneficial in low-data regimes}, where a jointly conditioned model lacks sufficient data to learn the correct modality weighting. FDP enforces conditioning on the most essential modality, leading to stronger overall performance. SmolVLA  fails drastically at \texttt{SweepToDustpan} (Figure \ref{fig:vla-dp3}) that requires precise spatial motions to sweep all the dirt particles into the dustpan. This highlights that there is potential for VLAs to improve beyond table-top tasks where algorithmic approaches like FDP do better. For M3L, we observe in Table \ref{tab:m3l_side} that the prioritization order of \texttt{vision>tactile} outperforms the joint-conditioning approach by over $20\%$ at $100$ and $200$ demonstrations. Several works  \cite{zhao2025touchbeginsvisionends,ankile2025imitation} have fine-tuned a BC policy using reinforcement learning, and FDP can serve as a performant prior policy for further fine-tuning. These results clearly show that FDP leads to more performant policies across various observational modalities, especially in low-data regimes.

\subsection{Robustness Gain Through Modality Prioritization.}
Prioritization prevents the model from developing spurious correlations by learning a residual policy for modalities with limited influence on robot actions. To test this, we evaluate FDP (\texttt{prop>vision}) against a jointly conditioned diffusion policy in environments with color variations and clutter. Both DP-DiT and FDP are trained on 100 demonstrations collected in the original environment and evaluated in three settings: the original, color-variant, and cluttered environments as shown in Figure \ref{fig:rlbench_dist}. \textbf{FDP significantly outperforms DP-DiT in both color-variant and cluttered settings by more than $40\%$}. We further collect five demonstrations in each modified environment to study few-shot adaptation to out-of-distribution data. FDP adapts more effectively, improving its performance by $15\%$ on average compared to $10\%$ for DP-DiT. This is achieved by updating only the residual model $\pi_\text{res}$ with new demonstrations, which adjusts the conditional distribution on visual modalities $p(y^{vis} \mid x,y^{prop})$ without modifying the full conditional action distribution $p(x \mid y^{prop},y^{vis})$. We extend this setting to point clouds (\texttt{pcd>vision}), where FDP learns a \texttt{vision} $\pi_\text{res}$ over DP3 used as $\pi_\text{base}$ \cite{ze20243d}, and compare it to DP3 with RGB inputs. Point clouds are sample-efficient for policy learning since they encode scene geometry in a single modality \cite{ze20243d}. However, our distractor experiments show that FDP with a visual residual over DP3 achieves ${\sim}20\%$ higher performance than DP3 using RGB inputs. These results clearly outline the robustness gain on adopting FDP as the policy formulation.

\begin{table}[t]
\centering
\caption{Block Pick success rates for 10 and 50 demonstrations.}
\label{tab:block-pick-st}
\begin{tabular}{lccc}
\toprule
\textbf{10 demos} & \textbf{S} & \textbf{M} & \textbf{L} \\
\midrule
DP-DiT              & 29.7 $\pm$ 3.1 & 12.0 $\pm$ 1.0 & 3.3 $\pm$ 0.6 \\
FDP (prop $>$ vision) & \textbf{73.7} $\pm$\textbf{ 3.8} & \textbf{21.3} $\pm$ \textbf{3.5} & 6.3 $\pm$ 3.1 \\
FDP (vision $>$ prop) & 18.7 $\pm$ 2.3 & 6.7 $\pm$ 1.2  & 3.3 $\pm$ 3.1 \\
\midrule
\textbf{50 demos} & \textbf{S} & \textbf{M} & \textbf{L} \\
\midrule
DP-DiT              & 95.3 $\pm$ 3.2 & \textbf{69.0} $\pm$ \textbf{7.0} & 45.7 $\pm$ 7.1 \\
FDP (prop $>$ vision) & \textbf{98.7} $\pm$ \textbf{1.5} & 55.0 $\pm$ 2.6 & 20.3 $\pm$ 3.5 \\
FDP (vision $>$ prop) & 96.7 $\pm$ 1.2 & \textbf{65.3} $\pm$ \textbf{8.1} & \textbf{60.0} $\pm$ \textbf{2.0} \\
prop $\pi_\text{base}$     & 47.3 $\pm$ 2.5 & 5.0 $\pm$ 1.7 & 1.3 $\pm$ 1.2 \\
vision $\pi_\text{base}$     & 96.0 $\pm$ 2.0 & \textbf{68.0} $\pm$ \textbf{6.0} & 51.3 $\pm$ 8.1 \\
\bottomrule
\end{tabular}
\end{table}

\section{Effects of Prioritization}
\label{sec:analysis}
\noindent \textbf{Intuition on the Order of Modality Prioritization}
As for most hyperparameters, intuition can be developed for the order of modality prioritization. To experimentally test this, we develop 3 variants of the block pick environments- \texttt{S}: $0.15\times0.2m$, \texttt{M}: $0.35\times0.45m$ and \texttt{L}: $0.55\times0.75m$, with increasing range of generalization in initial block-placement. The results are presented in Table \ref{tab:block-pick-st}. We observe that prioritization of proprioception out-performs all other models for the \texttt{S} environment while prioritization of vision tends to do better for larger \texttt{L} environment. This also conforms to the intuition of vision playing a diminished role when object placement area is smaller and the motions are repetitive, and a more significant role when the motion varies significantly based on the placement of the object. Tasks that correlate heavily with robot proprioception are not uncommon as the robot is solving them in the first person view, and can move close to the object if required. Our results in the M3L visuo-tactile environment also conform with \texttt{localize-then-execute} strategy \cite{zhao2025touchbeginsvisionends}, where the visual modality plays a more important role in localizing the hole for peg insertion. Learning a \texttt{tactile} $\pi_\text{res}$ over the \texttt{vision} $\pi_\text{base}$ learns residual scores for states where tactile influences the actions of the robot. Finally, the chosen action representation also plays a role, with FDP (\texttt{prop>}) realizing higher success rate for joint-state actions.

\noindent\textbf{Ablations.}  Table \ref{tab:door_ablation_10demos} presents ablations for our method. Results for DiT-Base show that lack of performance cannot be compensated for by increasing the model size. We also show that the integrated form of composition presented in Figure \ref{fig:archie}[c] outperforms the output score composition method [b] by a significant margin. Further, we find that preserving the diversity of $\pi_{\text{base}}$ is essential: overfitting the base model leaves little residual signal to learn, reducing generalization, while stopping the training too early leaves an unstable base. Our ablations show that selecting the $\pi_{\text{base}}$ checkpoint with the lowest validation loss (at 700 epochs) provides a good foundation for residual learning. Finally, the results in Table \ref{tab:block-pick-st} show that learning the residual is crucial. Policy performance with $\pi_\text{base}$ is unsatisfactory, and \textbf{our factored approach is able to improve the performance without diminishing the policy robustness given an additional modality}.  

\begin{table}[t]
\centering
\footnotesize
\caption{Ablation results on the Open Door task (10 demonstrations).}
\label{tab:door_ablation_10demos}
\begin{tabular}{@{}c!{\vrule width 0.4pt}c@{}}
\begin{tabular}{l|c}
\toprule
\textbf{Model} & \textbf{Succ. (\%)} \\
\midrule
DiT: small ($\sim$33M) & 24.0 {\scriptsize$\pm$ 7.2} \\
DiT: base ($\sim$130M) & 27.3 {\scriptsize$\pm$ 5.0} \\
Score Comp.: [b] Fig.~\ref{fig:archie} & 20.7 {\scriptsize$\pm$ 8.3} \\
\textbf{FDP [c]: Conv} & \textbf{42.0 {\scriptsize$\pm$ 5.2}} \\
\bottomrule
\end{tabular}
&
\begin{tabular}{l|c}
\toprule
\textbf{$\pi_\text{base}$ ep} & \textbf{Succ. (\%)} \\
\midrule
100 ep   & 24.7 {\scriptsize$\pm$ 6.1} \\
\textbf{700 ep} & \textbf{42.0 {\scriptsize$\pm$ 5.2}} \\
1500 ep  & 40.0 {\scriptsize$\pm$ 6.0} \\
2000 ep  & 40.7 {\scriptsize$\pm$ 3.1} \\
\bottomrule
\end{tabular}
\end{tabular}
\end{table}

\section{Real-world Experiments}
\begin{figure*}[t]
    \centering\includegraphics[height=4.5cm]{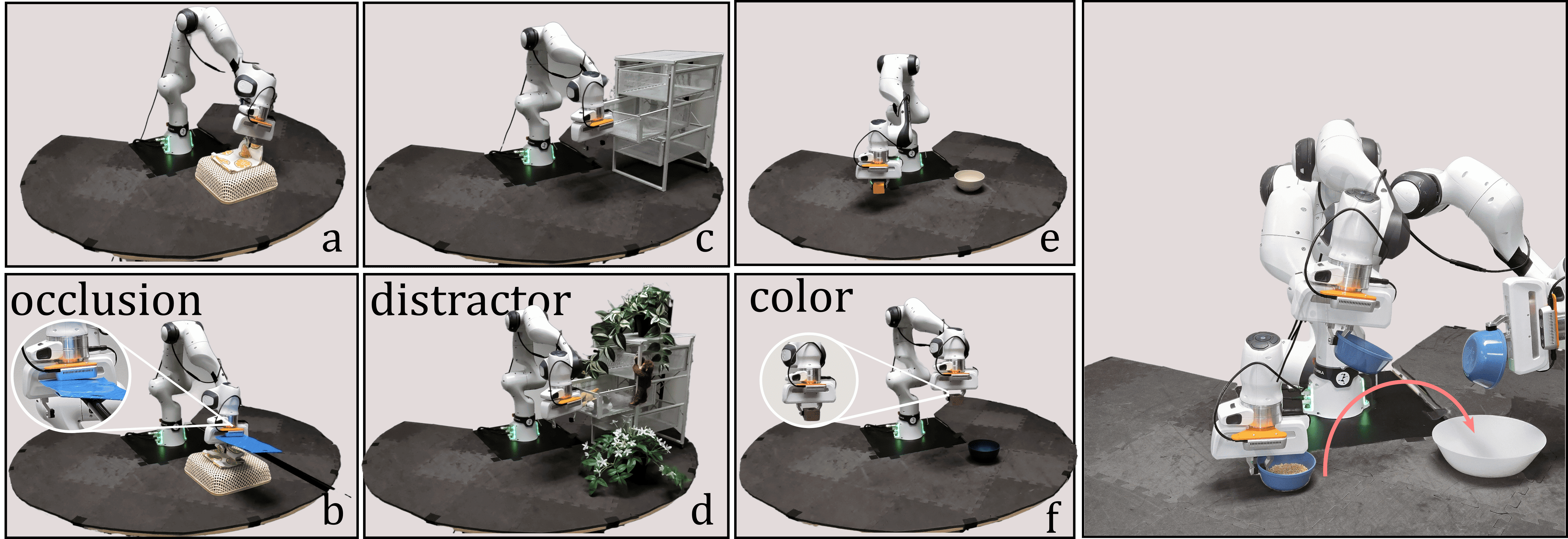}
    \caption{Real robot task domains and their variations. We show that DP-DiT fails in the presence of visual distribution shifts, highlighting the utility of prioritization for tasks with repetitive motions. Furthermore, FDP (\texttt{prop>vision}) is  robust to camera blinds which can cause safety risks for DP-DiT. }
    \label{fig:rr_dist}
\end{figure*}
We evaluate FDP and the DP-DiT baseline across four real-world domains and report their task success rates. The domains are -- \textit{Close Drawer} as a simple task where the robot has to push the drawer; \textit{Put Block in Bowl} that assesses the policy’s ability to perform precise pick-and-place actions; \textit{Pour in Bowl} to evaluate the policy’s dexterity in operating near joint limits and \textit{Fold Towel} to assess effectiveness in manipulating deformable objects.

We collect $50$ demonstrations per domain on a Franka FR3 robot using a $6$D space mouse, recording both proprioceptive and visual observations from two cameras—one mounted on the gripper and a static camera covering the workspace. The trained policies are evaluated on four task variations in each domain: (a) \texttt{default}: an in-distribution setup matching the conditions used during demonstration collection; (b) \texttt{color}: the object’s color is altered to test robustness to visual appearance changes; (c) \texttt{distractor}: novel, unseen objects such as vegetation props and soft toys are added to the scene to introduce clutter; and (d) \texttt{occlusion}: visual input is intermittently blocked during policy rollout to simulate partial observability. Figure \ref{fig:rr_dist} shows different task domains and their variations used in our experiments. We use $10$ rollouts in each experiment and report the task success rate as shown in Table \ref{table:real_world_results}. 

\begin{table}[htbp]
\centering
\caption{Success rates of DP-DiT and FDP across real-world tasks.}
\footnotesize
\setlength{\tabcolsep}{3pt}
\rowcolors{2}{white}{green!10}
\begin{tabular}{l|cc|cc|cc|cc}
\toprule
\textbf{Task Domain} 
& \multicolumn{2}{c|}{\texttt{default}} 
& \multicolumn{2}{c|}{\texttt{color}} 
& \multicolumn{2}{c|}{\texttt{dist.}} 
& \multicolumn{2}{c}{\texttt{occl.}} \\
& \textbf{DP} & \textbf{FDP} 
& \textbf{DP} & \textbf{FDP} 
& \textbf{DP} & \textbf{FDP} 
& \textbf{DP} & \textbf{FDP} \\
\midrule
Close Drawer       & \textbf{90} & \textbf{90} & \textbf{90} & \textbf{90} & 10 & \textbf{80} & 0 & \textbf{80} \\
Put Block in Bowl  & \textbf{80} & \textbf{80} & 0 & \textbf{60} & 0 & \textbf{60} & 10 & \textbf{60} \\
Pour in Bowl       & 70 & \textbf{80} & 40 & \textbf{80} & 20 & \textbf{60} & 10 & \textbf{50} \\
Fold Towel         & 40 & \textbf{60} & 40 & \textbf{70} & 30 & \textbf{70} & 10 & \textbf{50} \\
\bottomrule
\end{tabular}
\vspace{4pt}
\label{table:real_world_results}
\vspace{-0.5cm}
\end{table}

\noindent\textbf{Result Analysis.} We find that FDP is robust to distribution shifts in the environment. DP-DiT regularly produces unachievable robot actions under \texttt{distractor} and \texttt{occlusion} settings, often triggering safety stops, resulting in task failure. 
In contrast, FDP guided by its \texttt{prop} $\pi_{base}$, consistently generates stable actions even under severe occlusions and cluttered scenes, yielding an average absolute performance improvement of $40\%$. In the \texttt{default} experiment we observe that the FDP policy outperforms DP-DiT in the pouring and towel-folding tasks, which require precise object manipulation. \textbf{Notably, FDP achieves $5\times$ success-rate than that of DP-DiT in the camera occlusion setting, highlighting it's practicality for robots that must operate reliably in visually degraded environments.}

\section{Conclusion and Future Work}
We present Factorized Diffusion Policies (FDP), a novel policy formulation that factorizes the joint conditioning in diffusion models so that observational modalities can have a differing influence on the action diffusion process by design. We derive a novel loss function to realize the prioritization order of modalities and propose a novel architecture for efficient training. Through extensive experiments across visual, point-cloud, proprioceptive and tactile modalities, we demonstrate several benefits of modality prioritization, including improved sample efficiency and increased robustness. Overall, we observe $~15\%$ absolute performance improvement on more than 20 tasks spread across several observational modalities after adopting FDP over jointly conditioned diffusion policy and even SmolVLA. \textbf{We believe that this novel paradigm of modality prioritization along with strong performance gains, especially in low-data regimes make FDP a valuable contribution to the robot learning community}. Finally, our real-world experiments highlight the practical value of FDP in being a safe-to-deploy policy in the face of visual disturbances and even \textit{camera occlusions}, outperforming diffusion policies by over $40\%$.

FDP opens new avenues for future research, such as scalable integration of diversely sourced observational modality data for robot policy learning. FDP presents a computational overhead of searching through the prioritization order, and future work can develop a stronger framework that infers the modality to prioritize based on the task or the collected data. FDP prioritizes a single modality for the whole duration of the task, and dynamic prioritization of modalities also presents an interesting avenue for future work. Finally, we believe our approach will also have implications for fine-tuning of VLAs on new modalities not encountered in training. We believe FDP is a novel first-step towards the goal of observation modality prioritization. 

\bibliographystyle{IEEEtran}
\bibliography{references}  

\appendix
\subsection{Proof of Diffusion Loss for Full Conditional Action Distribution} \label{app:lma}
Most treatments of diffusion models have been studied primarily in the context of single-modality distributions, such as those over image pixels \cite{ho2020denoisingdiffusionprobabilisticmodels, song2022denoisingdiffusionimplicitmodels, song2020generative}. This formulation has been directly adopted by the robotics community \cite{chi2023diffusion,reuss2023goal,liu2025diffusion}, leading to the optimization objective shown in Equation \ref{eq:diff-loss-robo}.
\begingroup
\allowdisplaybreaks
\begin{align}
    \mathcal{L}_t(\vtheta) &= \mathbb{E}_{(\vx_0, \vy) \sim q(\vx_0, \vy),\, \vepsilon_0 \sim \mathcal{N}(0, \mathcal{I})}\left[||\vepsilon_0 -\hat{\vepsilon}_\vtheta(\vx_t,\vy,t)||_2^2\right] \label{eq:diff-loss-robo}
\end{align}
\endgroup
We formally show that a conditional diffusion process as defined by Dhariwal and Nichol \cite{dhariwal2021diffusion} results in Equation \ref{eq:diff-loss-robo} being a maximizer of the reweighted variational lower bound \cite{ho2020denoisingdiffusionprobabilisticmodels} on the log-likelihood of the conditional data distribution $\log q(\vx|\vy)$. This implies that by incorporating the observational modalities into the conditioning process, diffusion policies \cite{chi2023diffusion} learn the reverse transition kernel for action conditioned on all the modalities, as shown in the following proof. We adopt a slightly different notation from the main paper where we use $q$ for the forward and the reverse diffusion process, and the resulting marginals. 

\begin{lemma}
The diffusion loss function $\mathcal{L}_t(\vtheta)$ as defined in Equation \ref{eq:diff-loss-robo}, in expectation over the time-steps $1 \leq t \leq T$, maximizes the reweighted variational lower bound \cite{ho2020denoisingdiffusionprobabilisticmodels} on the log-likelihood of the conditional data distribution $\log q(\vx|\vy)$, under a Markovian noising process $\hat{q}(\vx_{t}|\vx_{t-1})$ and the conditional reverse transition kernel as $\hat{q}(\vx_{t-1}|\vx_{t}, \vy)$.
\end{lemma}

Here, we derive the diffusion loss function for the conditional distribution $p(\vx|\vy)$ instead of only $p(\vx)$. A parallel derivation for conditional variational auto-encoders can be found in Doersch \cite{doersch2016tutorial}. Following Dhariwal and Nichol \cite{dhariwal2021diffusion}, we start with a conditional Markovian noising forward process $\hat{q}$ similar to $q(\vx_t|\vx_{t-1}) = \mathcal{N}(\vx_t;\sqrt{\alpha_t}\vx_{t-1},(1-\alpha_t)\mathcal{I})$, and define the following:
\begingroup
\allowdisplaybreaks
\begin{align}
    \hat{q}(\vx_0) &:= q (\vx_0) \label{eq1}\\
    \hat{q}(\vx_{t+1}|\vx_t, \vy) &:= q(\vx_{t+1}|\vx_t) \label{eq2}\\
    \hat{q}(\vx_{1:T}|\vx_0, \vy) &:= \prod^T_{t=1}\hat{q}(\vx_{t}|\vx_{t-1}, \vy) \label{eq2b}
\end{align}
\endgroup
We now reproduce some results that will be used later in the derivation of diffusion loss for conditional distributions. Dhariwal and Nichol \cite{dhariwal2021diffusion} also show that
\begingroup
\allowdisplaybreaks
\begin{align}
    \scalemath{0.85}{\hat{q}(\vy|\vx_t,\vx_{t+1})} &= \scalemath{0.85}{\hat{q}(\vx_{t+1}|\vx_t,\vy)\frac{\hat{q}(\vy|\vx_t)}{\hat{q}(\vx_{t+1}|\vx_t)}} \\
     &= \scalemath{0.85}{\hat{q}(\vx_{t+1}|\vx_t)\frac{\hat{q}(\vy|\vx_t)}{\hat{q}(\vx_{t+1}|\vx_t)}} \\
     &= \scalemath{0.85}{\hat{q}(\vy|\vx_t)}
     \label{eq3}
\end{align}
\endgroup
Moreover, the unconditional reverse transition kernels can be shown to be equal using Bayes theorem, given Equations \ref{eq1} and \ref{eq2}: $\hat{\vq}(\vx_t|\vx_{t+1})=\vq(\vx_t|\vx_{t+1})$. Dhariwal and Nichol \cite{dhariwal2021diffusion} use the result from Equation \ref{eq3} to show the following for conditional reverse transition kernels.
\begingroup
\allowdisplaybreaks
\begin{align}
    \scalemath{0.85}{\hat{q}(\vx_t|\vx_{t+1},\vy)} &= \scalemath{0.85}{\frac{\hat{q}(\vx_t, \vx_{t+1},\vy)}{\hat{q}(\vx_{t+1},\vy)}} \\
                           &= \scalemath{0.85}{\frac{\hat{q}(\vx_t, \vx_{t+1},\vy)}{\hat{q}(\vy|\vx_{t+1})\hat{q}(\vx_{t+1})}} \\
                           &= \scalemath{0.85}{\frac{\hat{q}(\vx_t| \vx_{t+1})\hat{q}(\vy|\vx_t,\vx_{t+1})\hat{q}(\vx_{t+1})}{\hat{q}(\vy|\vx_{t+1})\hat{q}(\vx_{t+1})}} \\
                           &= \scalemath{0.85}{\frac{\hat{q}(\vx_t| \vx_{t+1})\hat{q}(\vy|\vx_t,\vx_{t+1})}{\hat{q}(\vy|\vx_{t+1})}} \\
                           &= \scalemath{0.85}{\frac{q(\vx_t| \vx_{t+1})\hat{q}(\vy|\vx_t)}{\hat{q}(\vy|\vx_{t+1})}}
    \label{eq4}
\end{align}
\endgroup
Further, we can show the following using Equations \ref{eq2} and \ref{eq2b} and the Markovian noising process. It states that the joint distribution of the noised samples conditioned on $\vy$ and $\vx_0$ are the same for both $\hat{q}$ and $q$.
\begingroup
\allowdisplaybreaks
\begin{align}
    \scalemath{0.85}{\hat{q}(\vx_{1:T}|\vx_0, \vy)} &= \scalemath{0.85}{\prod^T_{t=1}\hat{q}(\vx_{t}|\vx_{t-1}, \vy)} \\
    &= \scalemath{0.85}{\prod^T_{t=1}q(\vx_{t}|\vx_{t-1})} \\
    &= \scalemath{0.85}{q(\vx_{1:T}|\vx_0)} \label{eq:joint_dist}   
\end{align}
\endgroup
 We adapt the derivation of diffusion loss from Luo \cite{luo2022understandingdiffusionmodelsunified} to work with conditional distributions by maximizing the log-likelihood of the conditional data distribution $\log p(\vx|\vy)$ leading to evidence lower bound (ELBO).
\begingroup
\allowdisplaybreaks
\begin{align}
\scalemath{0.85}{\log p(\vx|\vy)}
&= \scalemath{0.85}{\log \int p(\vx_{0:T}|\vy) d\vx_{1:T}}\\
&= \scalemath{0.85}{\log \int \frac{p(\vx_{0:T}|\vy)\hat{q}(\vx_{1:T}|\vx_0,\vy)}{\hat{q}(\vx_{1:T}|\vx_0,\vy)} d\vx_{1:T}}\\
&= \scalemath{0.85}{\log \mathbb{E}_{\hat{q}(\vx_{1:T}|\vx_0,\vy)}\left[\frac{p(\vx_{0:T}|\vy)}{\hat{q}(\vx_{1:T}|\vx_0,\vy)}\right]}\\
&\geq \scalemath{0.85}{\mathbb{E}_{\hat{q}(\vx_{1:T}|\vx_0,\vy)}\left[\log \frac{p(\vx_{0:T}|\vy)}{\hat{q}(\vx_{1:T}|\vx_0,\vy)}\right]}
\end{align}
\endgroup
The ELBO can be further simplified as follows
\begingroup
\allowdisplaybreaks
\scalebox{0.85}{
\parbox{\linewidth}{
\begin{align}
\log p(\vx|\vy)
&\geq \mathbb{E}_{\hat{q}(\vx_{1:T}|\vx_0,\vy)}\left[\log \frac{p(\vx_{0:T}|\vy)}{\hat{q}(\vx_{1:T}|\vx_0,\vy)}\right] \notag \\
&= \mathbb{E}_{\hat{q}(\vx_{1:T}|\vx_0,\vy)}\left[\log \frac{p(\vx_T|\vy)\prod_{t=1}^{T}p_{\bm{\theta}}(\vx_{t-1}|\vx_t,\vy)}{\prod_{t = 1}^{T}\hat{q}(\vx_{t}|\vx_{t-1},\vy)}\right] \notag \\
&= \mathbb{E}_{\hat{q}(\vx_{1:T}|\vx_0,\vy)}\left[
\log \frac{p_(\vx_T|\vy)p_{\bm{\theta}}(\vx_0|\vx_1,\vy)}{\hat{q}(\vx_1|\vx_0,\vy)} \right. \notag \\
&\quad \left. + \log \prod_{t=2}^{T}\frac{p_{\bm{\theta}}(\vx_{t-1}|\vx_t,\vy)}{\hat{q}(\vx_{t}|\vx_{t-1}, \vx_0,\vy)}\right] \notag \\
&= \mathbb{E}_{\hat{q}(\vx_{1:T}|\vx_0,\vy)}\left[
\log \frac{p(\vx_T|\vy)p_{\bm{\theta}}(\vx_0|\vx_1,\vy)}{\hat{q}(\vx_1|\vx_0,\vy)} \right. \notag \\
&\quad \left. + \log \prod_{t=2}^{T}\frac{p_{\bm{\theta}}(\vx_{t-1}|\vx_t,\vy)}{
\frac{\hat{q}(\vx_{t-1}|\vx_{t}, \vx_0,\vy)\hat{q}(\vx_t|\vx_0,\vy)}{\hat{q}(\vx_{t-1}|\vx_0,\vy)}
}\right] \notag \\
&= \mathbb{E}_{\hat{q}(\vx_{1:T}|\vx_0,\vy)}\left[
\log \frac{p(\vx_T|\vy)p_{\bm{\theta}}(\vx_0|\vx_1,\vy)}{\hat{q}(\vx_1|\vx_0,\vy)} \right. \notag \\
&\quad \left. + \log \frac{\hat{q}(\vx_1|\vx_0,\vy)}{\hat{q}(\vx_T|\vx_0,\vy)} 
+ \log \prod_{t=2}^{T}\frac{p_{\bm{\theta}}(\vx_{t-1}|\vx_t,\vy)}{\hat{q}(\vx_{t-1}|\vx_{t}, \vx_0,\vy)}\right] \notag \\
&= \mathbb{E}_{\hat{q}(\vx_{1:T}|\vx_0,\vy)}\left[
\log \frac{p(\vx_T|\vy)p_{\bm{\theta}}(\vx_0|\vx_1,\vy)}{\hat{q}(\vx_T|\vx_0,\vy)} \right. \notag \\
&\quad \left. +  \sum_{t=2}^{T}\log\frac{p_{\bm{\theta}}(\vx_{t-1}|\vx_t,\vy)}{\hat{q}(\vx_{t-1}|\vx_{t}, \vx_0,\vy)}\right] \notag \\
&= \mathbb{E}_{\hat{q}(\vx_{1}|\vx_0,\vy)}\left[\log p_{\bm{\theta}}(\vx_0|\vx_1,\vy)\right] \notag \\
&\quad + \mathbb{E}_{\hat{q}(\vx_{T}|\vx_0,\vy)}\left[\log \frac{p(\vx_T|\vy)}{\hat{q}(\vx_T|\vx_0,\vy)}\right] \notag \\
&\quad + \sum_{t=2}^{T}\mathbb{E}_{\hat{q}(\vx_{t}, \vx_{t-1}|\vx_0,\vy)}\left[
\log\frac{p_{\bm{\theta}}(\vx_{t-1}|\vx_t,\vy)}{\hat{q}(\vx_{t-1}|\vx_{t}, \vx_0,\vy)}\right] \notag \\
&= \underbrace{\mathbb{E}_{\hat{q}(\vx_{1}|\vx_0,\vy)}\left[\log p_{\bm{\theta}}(\vx_0|\vx_1,\vy)\right]}_\text{reconstruction term} \notag \\
&\quad - \underbrace{\infdiv{\hat{q}(\vx_T|\vx_0,\vy)}{p(\vx_T|\vy)}}_\text{prior matching term} \notag \\
&\quad - \sum_{t=2}^{T} \underbrace{
\mathbb{E}_{\hat{q}(\vx_{t}|\vx_0,\vy)}\left[
\right.
}_\text{denoising matching term} \notag \\
&\hspace{3em} \underbrace{
\left.
\infdiv{\hat{q}(\vx_{t-1}|\vx_t, \vx_0,\vy)}{p_{\bm{\theta}}(\vx_{t-1}|\vx_t,\vy)}
\right]
}_{}
\label{eq:6}
\end{align}
}}
\endgroup

\newpage
The reconstruction term is ignored for training \cite{ho2020denoisingdiffusionprobabilisticmodels,luo2022understandingdiffusionmodelsunified}, and the prior matching term does not have any trainable parameters. We further simplify the denoising matching term using Equation \ref{eq4} further conditioned on $\vx_0$.
\begingroup
\allowdisplaybreaks
\scalebox{0.85}{
\parbox{\linewidth}{
\begin{align}
& -\sum_{t=2}^{T} \mathbb{E}_{\hat{q}(\vx_{t}|\vx_0,\vy)}\left[
\infdiv{\hat{q}(\vx_{t-1}|\vx_t, \vx_0,\vy)}{
p_{\bm{\theta}}(\vx_{t-1}|\vx_t,\vy)}
\right] \notag \\
&= -\sum_{t=2}^{T} \mathbb{E}_{\hat{q}(\vx_{t}|\vx_0,\vy)}\left[
\mathbb{E}_{\hat{q}(\vx_{t-1}|\vx_{t},\vx_0,\vy)} \left[
\log \hat{q}(\vx_{t-1}|\vx_{t},\vx_0,\vy) \right. \right. \notag \\
&\left. \left. \hspace{1em}
- \log p_{\bm{\theta}}(\vx_{t-1}|\vx_t,\vy)
\right] \right] \notag \\
&= -\sum_{t=2}^{T} \mathbb{E}_{\hat{q}(\vx_{t}|\vx_0,\vy)}\left[
\mathbb{E}_{\hat{q}(\vx_{t-1}|\vx_{t},\vx_0,\vy)} \left[
\log \hat{q}(\vx_{t-1}|\vx_{t},\vx_0) \right. \right. \notag \\
&\left. \left. \hspace{1em}
+ \log \frac{\hat{q}(\vy|\vx_{t-1},\vx_0)}{\hat{q}(\vy|\vx_{t},\vx_0)}
- \log p_{\bm{\theta}}(\vx_{t-1}|\vx_t,\vy)
\right] \right] \notag \\
&= -\sum_{t=2}^{T} \mathbb{E}_{\hat{q}(\vx_{t}|\vx_0,\vy)}\left[
\infdiv{\hat{q}(\vx_{t-1}|\vx_t, \vx_0)}{
p_{\bm{\theta}}(\vx_{t-1}|\vx_t,\vy)}
\right] \notag \\
&\quad - \sum_{t=2}^{T} \mathbb{E}_{\hat{q}(\vx_{t}|\vx_0,\vy)}\left[
\mathbb{E}_{\hat{q}(\vx_{t-1}|\vx_{t},\vx_0,\vy)} \left[
\log \frac{\hat{q}(\vy|\vx_{t-1},\vx_0)}{\hat{q}(\vy|\vx_{t},\vx_0)}
\right] \right]
\label{eq:7}
\end{align}
}}
\endgroup

Note that the expectation is taken over a distribution independent of $\vy$, since $\hat{q}(\vx_{1:T}|\vx_0, \vy) = q(\vx_{1:T}|\vx_0) $, as shown in Equation \ref{eq:joint_dist}. It is easy to see that the first term in the resulting expression is the KL divergence between the model parameterized with the condition $\vy$ and the unconditional reverse transition kernel, leading to the popularly used diffusion loss of Equation \ref{eq:diff-loss-robo}. However, an additional term is introduced for the conditional diffusion process. This minimizes the difference in the likelihood of the labels between consecutive denoising steps. However, since it does not have trainable parameters, we ignore it. 

In Equation \ref{eq:diff-loss-robo}, $\vepsilon_\vtheta(\vx_t,\vy,t)$ arises from the reparametrization of the reverse transition kernel $q_{\vtheta}(\vx_{t-1}|\vx_{t}, \vy^{1:M})$. In this work, we argue that learning the full conditional directly is restrictive in several aspects of robot learning. Firstly, it necessitates the joint collection of the robot action and all observational modalities. Secondly, the model is vulnerable to even small distribution shifts in \textit{any} modality. These shifts  require a prohibitively large amount of data to address when the observation modalities are high-dimensional. Finally, among the multiple observation modalities it is hard to pinpoint the level of each mode's task dependent influence with limited data. 

\newpage
\subsection{Proof for FDP Loss}
\label{app:thm-proof}
\begin{theorem}
    \label{app:thm:1}
    Explicit score matching for $\vs_\vphi(\vx_t, \vy^{1:M})$ in Equation \ref{eq:esm} is equivalent to the objective $\mathcal{J}^{res}_{\alpha_t}(\vphi)$:
    \begin{equation}
    \label{app:eq:codig-loss}
    \scalebox{1}{
    \parbox{\linewidth}{
    \begin{align*}
        \mathbb{E}_{\stackrel{p_{\alpha_t}(\vx, \vx_t,}{\vy^{1:M})}}
        \Biggl[
        \frac{1}{2} \left\| \nabla_{\vx_t} \log p_{\alpha_t}(\vx_t|\vx) {-}
        \begin{array}{l}
        \vs^*(\vx_t,\vy^{1:k}) \\
        - \vs_{\vphi}(\vx_t, \vy^{1:M})
        \end{array}
        \right\|^2_2 \Biggr]
    \end{align*}
    }}
    \end{equation}
\end{theorem}
Here $\vs^*(\vx_t,\vy^{1:k})$ is the the frozen optimal score model for $\nabla_{\vx_t} \log p(\vx_t|\vy^{1:k})$, approximated using a learned model $\pi_\text{base}:\vs_{\vtheta}(\vx_t,\vy^{1:k})$ in practice.

Chao et al. \cite{chao2022denoisinglikelihoodscorematching} in their insightful work for score-based models, show that the following two losses differ only by a constant, where the $\tilde{\vx}$ and $\tilde{\vy}$ notation is introduced to indicate Gaussian noised variables, with variances $\alpha$ and $\tau$ respectively.

\begingroup
\allowdisplaybreaks
\scalebox{0.75}{
\parbox{\linewidth}{
\begin{align}
&\mathcal{D}_F(p_\vphi(\tilde{\vy}|\tilde{\vx}) \| p_{\alpha, \tau}(\tilde{\vy}|\tilde{\vx}))
= \mathbb{E}_{p_{\alpha, \tau}(\tilde{\vx},\tilde{\vy})} \Biggl[
\frac{1}{2} \left\|
\begin{array}{l}
\nabla_{\tilde{\vx}} \log p_\vphi(\tilde{\vy}|\tilde{\vx}) \\
- \nabla_{\tilde{\vx}} \log p_{\alpha, \tau}(\tilde{\vy}|\tilde{\vx})
\end{array}
\right\|^2_2 \Biggr] \\
&\mathcal{L}_{DLSM}(\vphi) 
= \mathbb{E}_{p_{\alpha, \tau}(\vx,\tilde{\vx}, \vy, \tilde{\vy})} \Biggl[
\frac{1}{2} \left\|
\begin{array}{l}
\nabla_{\tilde{\vx}} \log p_\vphi(\tilde{\vy}|\tilde{\vx}) 
+ \nabla_{\tilde{\vx}} \log p_\vtheta(\tilde{\vx}) \\
- \nabla_{\tilde{\vx}} \log p_\alpha(\tilde{\vx}|\vx)
\end{array}
\right\|^2_2 \Biggr]
\end{align}
}}
\endgroup

We extend their proof for diffusion models and multiple conditionals below. Here $\vx_t$ is noised with the forward transition kernel $p_{\bar{\alpha}_t}(\vx_t|\vx){=}\mathcal{N}(\vx_t;\sqrt{\Bar{\alpha_t}}\vx,(1{-}\Bar{\alpha_t})I)$. Explicit Score Matching loss between the model and the true score of the classifier can be further expanded as:
\begingroup
\allowdisplaybreaks
\begin{align}
    &\scalemath{0.85}{D^t_F(p_\vphi(\tilde{\vy}^{k+1:M}|\vx_t,\tilde{\vy}^{1:k})||p_{\alpha, \tau}
    (\tilde{\vy}^{k+1:M}|\vx_t,\tilde{\vy}^{1:k}))} \\
    &= \scalemath{0.85}{\mathbb{E}_{p_{\alpha, \tau}(\vx_t, \tilde{\vy}^{1:M})}\left[
    \frac{1}{2}||\nabla_{\vx_t} \log p_\vphi(\tilde{\vy}^{k+1:M}|\vx_t,\tilde{\vy}^{1:k})\right.} \notag \\
    &\qquad\scalemath{0.85}{\left. - \nabla_{\vx_t} \log p_{\alpha, \tau}(\tilde{\vy}^{k+1:M}|\vx_t,
    \tilde{\vy}^{1:k})||^2_2\right]} \\
    &= \scalemath{0.85}{\mathbb{E}_{p_{\alpha, \tau}(\vx_t, \tilde{\vy}^{1:M})}\left[
    \frac{1}{2}||\nabla_{\vx_t} \log p_\vphi(\tilde{\vy}^{k+1:M}|\vx_t,\tilde{\vy}^{1:k})||^2_2\right]} \notag \\
    &\quad + \scalemath{0.85}{\mathbb{E}_{p_{\alpha, \tau}(\vx_t, \tilde{\vy}^{1:M})}\left[
    \frac{1}{2}||\nabla_{\vx_t} \log p_{\alpha, \tau}(\tilde{\vy}^{k+1:M}|\vx_t,\tilde{\vy}^{1:k})||^2_2\right]} \notag \\
    &\quad - \scalemath{0.85}{\mathbb{E}_{p_{\alpha, \tau}(\vx_t, \tilde{\vy}^{1:M})}\left[
    \left\langle\nabla_{\vx_t} \log p_\vphi(\tilde{\vy}^{k+1:M}|\vx_t,\tilde{\vy}^{1:k}), \right.\right.} \notag \\
    &\qquad\quad\scalemath{0.85}{\left.\left. \nabla_{\vx_t} \log p_{\alpha, \tau}(\tilde{\vy}^{k+1:M}
    |\vx_t,\tilde{\vy}^{1:k}) \right\rangle \right]} \\
    &= \scalemath{0.85}{\mathbb{E}_{p_{\alpha, \tau}(\vx_t, \tilde{\vy}^{1:M})}\left[
    \frac{1}{2}||\nabla_{\vx_t} \log p_\vphi(\tilde{\vy}^{k+1:M}|\vx_t,\tilde{\vy}^{1:k})||^2_2\right]} \notag \\
    &\quad + \scalemath{0.85}{\mathbb{E}_{p_{\alpha, \tau}(\vx_t, \tilde{\vy}^{1:M})}\left[
    \frac{1}{2}||\nabla_{\vx_t} \log p_{\alpha, \tau}(\tilde{\vy}^{k+1:M}|\vx_t,\tilde{\vy}^{1:k})||^2_2\right]} \notag \\
    &\quad - \scalemath{0.85}{\mathbb{E}_{p_{\alpha, \tau}(\vx_t, \tilde{\vy}^{1:M})}\left[
    \left\langle\nabla_{\vx_t} \log p_\vphi(\tilde{\vy}^{k+1:M}|\vx_t,\tilde{\vy}^{1:k}), \right.\right.} \notag \\
    &\qquad\quad\scalemath{0.85}{\left.\left. \nabla_{\vx_t} \log p_{\alpha, \tau}
    (\vx_t|\tilde{\vy}^{1:k},\tilde{\vy}^{k+1:M}) \right\rangle \right.} \notag \\
    &\qquad\quad\scalemath{0.85}{\left. \left. - \nabla_{\vx_t} \log p_{\alpha, \tau}
    (\vx_t|\tilde{\vy}^{1:k}) \right\rangle \right]} \\
    &= \scalemath{0.85}{\mathbb{E}_{p_{\alpha, \tau}(\vx_t, \tilde{\vy}^{1:M})}\left[
    \frac{1}{2}||\nabla_{\vx_t} \log p_\vphi(\tilde{\vy}^{k+1:M}|\vx_t,\tilde{\vy}^{1:k})||^2_2\right]} \notag \\
    &\quad + \scalemath{0.85}{\mathbb{E}_{p_{\alpha, \tau}(\vx_t, \tilde{\vy}^{1:M})}\left[
    \frac{1}{2}||\nabla_{\vx_t} \log p_{\alpha, \tau}(\tilde{\vy}^{k+1:M}|\vx_t,\tilde{\vy}^{1:k})||^2_2\right]} \notag \\
    &\quad + \scalemath{0.85}{\mathbb{E}_{p_{\alpha, \tau}(\vx_t, \tilde{\vy}^{1:M})}\left[
    \left\langle\nabla_{\vx_t} \log p_\vphi(\tilde{\vy}^{k+1:M}|\vx_t,\tilde{\vy}^{1:k}), \right. \right.} \notag \\
    &\qquad\quad\scalemath{0.85}{\left. \left. \nabla_{\vx_t} \log p_{\alpha, \tau}
    (\vx_t|\tilde{\vy}^{1:k}) \right\rangle \right]} \notag \\
    &\quad \underbrace{- \scalemath{0.85}{\mathbb{E}_{p_{\alpha, \tau}(\vx_t, \tilde{\vy}^{1:M})}\left[
    \left\langle\nabla_{\vx_t} \log p_\vphi(\tilde{\vy}^{k+1:M}|\vx_t,\tilde{\vy}^{1:k}), \right. \right.}}_{\text{Term 1}} \notag \\
    &\qquad\quad\scalemath{0.85}{\left. \left. \nabla_{\vx_t} \log p_{\alpha, \tau}
    (\vx_t|\tilde{\vy}^{1:k},\tilde{\vy}^{k+1:M}) \right\rangle \right]} \label{eq:term1}
\end{align}
\endgroup

Simplifying the Term 1 further:
\begingroup
\allowdisplaybreaks
\begin{align}
    &-\scalemath{0.85}{\mathbb{E}_{p_{\alpha, \tau}(\vx_t, \tilde{\vy}^{1:M})}\big[\langle\nabla_{\vx_t} \log p_\vphi(\tilde{\vy}^{k+1:M}|\vx_t,\tilde{\vy}^{1:k}),} \notag \\
    &\quad\scalemath{0.85}{\nabla_{\vx_t} \log p_{\alpha, \tau}(\vx_t|\tilde{\vy}^{1:M})\rangle\big]}  \notag \\
    &=- \scalemath{0.85}{\int_{\vx_{t}}\int_{\tilde{\vy}^{1:M}} p_{\tau}(\tilde{\vy}^{1:M})p_{\alpha,\tau}(\vx_t|\tilde{\vy}^{1:M}) } \notag \\
    &\quad\scalemath{0.85}{\langle\nabla_{\vx_t} \log p_\vphi(\tilde{\vy}^{k+1:M}|\vx_t,\tilde{\vy}^{1:k}), 
    \frac{\nabla_{\vx_t} p_{\alpha, \tau}(\vx_t|\tilde{\vy}^{1:M})}{p_{\alpha, \tau}(\vx_t|\tilde{\vy}^{1:M})}\rangle d\tilde{\vy}^{1:M} d\vx_t} \notag \\
    &=-\scalemath{0.85}{\int_{\vx_{t}}\int_{\tilde{\vy}^{1:M}} p_{\tau}(\tilde{\vy}^{1:M})
    \langle\nabla_{\vx_t} \log p_\vphi(\tilde{\vy}^{k+1:M}|\vx_t,\tilde{\vy}^{1:k}),} \notag\\
    &\quad\qquad\scalemath{0.85}{\nabla_{\vx_t} \int_{\vx_0} p_{0, \tau}(\vx_0|\tilde{\vy}^{1:M}) p_{\alpha, \tau}(\vx_t|\vx_0,\tilde{\vy}^{1:M})d\vx_0 \rangle d\tilde{\vy}^{1:M} d\vx_t} \notag \\
    &=- \scalemath{0.85}{\int_{\vx_{t}}\int_{\tilde{\vy}^{1:M}} p_{\tau}(\tilde{\vy}^{1:M})
    \langle\nabla_{\vx_t} \log p_\vphi(\tilde{\vy}^{k+1:M}|\vx_t,\tilde{\vy}^{1:k}),} \notag \\
    &\qquad\qquad\scalemath{0.85}{\nabla_{\vx_t} \int_{\vx_0}\int_{\vy^{1:M}} p_{0, \tau}(\vx_0|\tilde{\vy}^{1:M}) p_{\alpha, \tau}(\vx_t|\vx_0,\tilde{\vy}^{1:M},\vy^{1:M})\cdot } \notag \\
    &\qquad\qquad\scalemath{0.85}{p_(\vy^{1:M}|\vx_0,\tilde{\vy}^{1:M}) d\vy^{1:M} d\vx_0 \rangle d\tilde{\vy}^{1:M} d\vx_t} \notag \\
    &=- \scalemath{0.85}{\int_{\vx_{t}}\int_{\tilde{\vy}^{1:M}}\int_{\vx_0} \int_{\vy^{1:M}} p_{\tau}(\vx_0, \vx_t, \vy^{1:M}, \tilde{\vy}^{1:M})} \notag \\
    &\qquad\qquad\scalemath{0.85}{
    \langle\nabla_{\vx_t} \log p_\vphi(\tilde{\vy}^{k+1:M}|\vx_t,\tilde{\vy}^{1:k}),} \notag \\
    &\qquad\qquad\scalemath{0.85}{\nabla_{\vx_t} \log p_{\alpha, \tau}(\vx_t|\vx_0,\tilde{\vy}^{1:M},\vy^{1:M}) \rangle d\vy^{1:M} d\vx_0 d\tilde{\vy}^{1:M} d\vx_t} \notag \\
    &= -\scalemath{0.85}{\mathbb{E}_{p_{\alpha, \tau}(\vx_0, \vx_t, \vy^{1:M}, \tilde{\vy}^{1:M})}\big[\langle\nabla_{\vx_t} \log p_\vphi(\tilde{\vy}^{k+1:M}|\vx_t,\tilde{\vy}^{1:k}), } \notag \\
    &\qquad\qquad\scalemath{0.85}{\nabla_{\vx_t} \log p_{\alpha}(\vx_t|\vx_0) \rangle\big]} \notag
\end{align}
\endgroup

Plugging this back into Equation \ref{eq:term1}, we get-
\begingroup
\allowdisplaybreaks
\begin{align}
    &\scalemath{0.85}{\mathcal{D}_F\left(
    p_\vphi(\tilde{\vy}^{k+1:M}|\vx_t,\tilde{\vy}^{1:k})\middle\|
    p_{\alpha, \tau}(\tilde{\vy}^{k+1:M}|\vx_t,\tilde{\vy}^{1:k})
    \right)} \notag \\
    &= \scalemath{0.85}{\mathbb{E}_{p_{\alpha, \tau}(\vx_t, \tilde{\vy}^{1:M})}
    \left[\frac{1}{2} \left\| \nabla_{\vx_t} \log p_\vphi(
    \tilde{\vy}^{k+1:M}|\vx_t,\tilde{\vy}^{1:k}) \right\|_2^2 \right]} \notag \\
    &\quad + \scalemath{0.85}{\mathbb{E}_{p_{\alpha, \tau}(\vx_t, \tilde{\vy}^{1:M})}
    \left[\frac{1}{2} \left\| \nabla_{\vx_t} \log p_{\alpha, \tau}(
    \tilde{\vy}^{k+1:M}|\vx_t,\tilde{\vy}^{1:k}) \right\|_2^2 \right]} \notag \\ 
    &\quad + \scalemath{0.85}{\mathbb{E}_{p_{\alpha, \tau}(\vx_t, \tilde{\vy}^{1:M})}
    \left[\bigg\langle \nabla_{\vx_t} \log p_\vphi(
    \tilde{\vy}^{k+1:M}|\vx_t,\tilde{\vy}^{1:k}), \right.} \notag \\
    &\qquad\qquad\scalemath{0.85}{\left. \nabla_{\vx_t} \log p_{\alpha, \tau}(
    \vx_t|\tilde{\vy}^{1:k}) \bigg\rangle \right]} \notag \\
    &\quad - \scalemath{0.85}{\mathbb{E}_{p_{\alpha, \tau}(\vx_0, \vx_t, \vy^{1:M}, \tilde{\vy}^{1:M})}
    \left[\bigg\langle \nabla_{\vx_t} \log p_\vphi(
    \tilde{\vy}^{k+1:M}|\vx_t,\tilde{\vy}^{1:k}), \right.} \notag \\
    &\qquad\qquad\scalemath{0.85}{\left. \nabla_{\vx_t} \log p_{\alpha}(
    \vx_t|\vx_0) \bigg\rangle \right]} \label{eq:proof-step}
\end{align}
\endgroup

Here, $ \scalemath{0.85}{\mathbb{E}_{p_{\alpha, \tau}(\vx_t, \tilde{\vy}^{1:M})}\left[\frac{1}{2}||\nabla_{\vx_t} \log p_{\alpha, \tau}(\tilde{\vy}^{k+1:M}|\vx_t,\tilde{\vy}^{1:k})||^2_2\right]}$ is a constant. Further, adding the constant $\scalemath{0.85}{\mathbb{E}_{p_{\alpha, \tau}(\vx_t, \tilde{\vy}^{1:k})}\left[\frac{1}{2}||\nabla_{\vx_t} \log p_{\alpha, \tau}(\vx_t|\tilde{\vy}^{1:k}) - \nabla_{\vx_t} \log p_{\alpha}(\vx_t|\vx_0)||^2_2\right]}$ to Equation \ref{eq:proof-step}, we get:

\begingroup
\allowdisplaybreaks
\begin{align}
    &\scalemath{0.85}{\mathcal{D}_F\left(
    p_\vphi(\tilde{\vy}^{k+1:M}|\vx_t,\tilde{\vy}^{1:k}) \,\middle\|\, 
    p_{\alpha, \tau}(\tilde{\vy}^{k+1:M}|\vx_t,\tilde{\vy}^{1:k})
    \right)} \notag \\
    &= \scalemath{0.85}{\mathbb{E}_{p_{\alpha, \tau}(\vx_t, \tilde{\vy}^{1:M})}
    \left[\frac{1}{2} \left\| \nabla_{\vx_t} \log p_\vphi(\tilde{\vy}^{k+1:M}
    |\vx_t,\tilde{\vy}^{1:k}) \right\|_2^2 \right]} \notag \\
    &\quad + \scalemath{0.85}{\mathbb{E}_{p_{\alpha, \tau}(\vx_t, \tilde{\vy}^{1:M})}
    \left[\bigg\langle \nabla_{\vx_t} \log p_\vphi(\tilde{\vy}^{k+1:M}
    |\vx_t,\tilde{\vy}^{1:k}), \right.} \notag \\
    &\qquad\qquad\scalemath{0.85}{\left. \nabla_{\vx_t} \log p_{\alpha, \tau}
    (\vx_t|\tilde{\vy}^{1:k}) \bigg\rangle \right]} \notag \\
    &\quad - \scalemath{0.85}{\mathbb{E}_{p_{\alpha, \tau}(\vx_0, \vx_t, \vy^{1:M}, \tilde{\vy}^{1:M})}
    \left[\bigg\langle \nabla_{\vx_t} \log p_\vphi(\tilde{\vy}^{k+1:M}
    |\vx_t,\tilde{\vy}^{1:k}), \right.} \notag \\
    &\qquad\qquad\scalemath{0.85}{\left. \nabla_{\vx_t} \log p_{\alpha}
    (\vx_t|\vx_0) \bigg\rangle \right]} \notag \\
    &\quad + \scalemath{0.85}{\mathbb{E}_{p_{\alpha, \tau}(\vx_t, \tilde{\vy}^{1:k})}
    \bigg[\frac{1}{2} \left\| \nabla_{\vx_t} \log p_{\alpha, \tau}(\vx_t|\tilde{\vy}^{1:k}) \right.} \notag \\
    &\qquad\qquad\scalemath{0.85}{\left. - \nabla_{\vx_t} \log p_{\alpha}(\vx_t|\vx_0) \right\|_2^2 \bigg] + C} \\
    &= \scalemath{0.85}{\mathbb{E}_{p_{\alpha, \tau}(\vx,\vx_t, \vy^{1:M}, \tilde{\vy}^{1:M})}
    \biggl[ \frac{1}{2} \Big\| \nabla_{\vx_t} \log p_\vphi(\tilde{\vy}^{k+1:M}
    |\vx_t,\tilde{\vy}^{1:k})} \notag \\
    &\qquad\qquad\scalemath{0.85}{+ \nabla_{\vx_t} \log p_\vtheta(\vx_t|\tilde{\vy}^{1:k})
    - \nabla_{\vx_t} \log p_\alpha(\vx_t|\vx) \Big\|_2^2 \biggr] + C} \label{eq:proved}
\end{align}
\endgroup

Simplifying $\nabla_{\vx_t} \log p_\alpha(\vx_t|\vx)$ to $-\vepsilon_0/\sqrt{1-\bar{\alpha_t}}$, where $\vepsilon_0 \sim \mathcal{N}(0, \mathcal{I})$, and taking the $(1-\bar{\alpha_t})$ times the DSM objective \cite{song2020score}, we obtain the simplified diffusion loss:
\begingroup
\allowdisplaybreaks
\begin{align}
    \scalemath{0.85}{\mathcal{L}^t_{res}(\vphi)} 
    &= \scalemath{0.85}{\mathbb{E}_{p_{\tau}(\vx, \vy^{1:M}, \tilde{\vy}^{1:M})}
    \mathbb{E}_{\vepsilon_0 \sim \mathcal{N}(0, \mathcal{I})}
    \biggl[\Big\| \vepsilon_0} \notag \\
    &\qquad\scalemath{0.85}{
    - \vepsilon_{\theta}(\vx_t,\tilde{\vy}^{1:k}, t)
    + \hat{\vepsilon}_{\vphi}(\tilde{\vy}^{k+1:M}, \vx_t,\tilde{\vy}^{1:k}, t)
    \Big\|^2_2 \biggr]} + C \label{eq-app:fdp}
\end{align}
\endgroup

\subsection{Architecture and Implementation Details}
\label{app:archie}
All transformer-based models are trained over 2000 epochs for visual tasks and 3000 epochs for low-dimensional tasks. Unet \cite{ronneberger2015u} is trained over 3000 epochs for visual tasks and 5000 epochs for low-dimensional tasks. We train models on visual tasks with a batch size of 64, and low-dimensional tasks with a batch size of 256. All models are trained on NVIDIA A5000 or A40 GPUs, with training times ranging from 6 to 12 hours depending on model size and the number of camera inputs. Our current implementations support an action prediction latency of $\sim$50ms for DP-DiT, $\sim$100ms for UNet \cite{chi2023diffusion} and output composition of models as shown in Figure \ref{fig:archie} $[b]$ and $\sim$150ms for FDP model shown in $[c]$.

\textbf{DP-DiT.} We use DiT-S (${\sim}33$M parameters)\cite{peebles2023scalablediffusionmodelstransformers} as the base architecture, with $12$ layers, $6$ heads and a hidden dimension of $6$. Peebles et al. \cite{peebles2023scalablediffusionmodelstransformers} specifically show that the conditioning using AdaLn-Zero outperforms other forms of conditioning such as in-context and cross-attention for image generation. However, we observe slightly stronger performance when the weights for the final layer of AdaLn are initialized with a Gaussian. We use different untrained ResNet-18 (${\sim}12$M parameters)\cite{he2016deep} encoders for each camera and also encode proprioception using a separate encoder. All encoded conditionals across the observation horizon are concatenated before using AdaLn. The model size conditioned on the input images from 2 cameras is ${\sim}56$ M parameters. All DiT models are trained using a learning rate of $1e{-}4$ and a weight decay of $1e{-}3$. We also perform exponential moving average (EMA) to reduce the variance in training. The same DiT backbone is used for low-dimensional, visual, and point-cloud tasks. We use the 100-step DDPM \cite{ho2020denoisingdiffusionprobabilisticmodels} noise scheduler suggested by CHi et al. \cite{chi2023diffusion} implemented using HuggingFace Diffusers \cite{von-platen-etal-2022-diffusers}. Sampling is performed using 8-step DDIM \cite{song2022denoisingdiffusionimplicitmodels}.

\textbf{DP-UNet.} We use the 1D-UNet \cite{ronneberger2015u} implementation from Chi et al. \cite{chi2023diffusion}. UNet is trained using a learning rate of $1e{-}4$ and a weight decay of $1e{-}6$. Although the parameter count of the DiT model does not vary significantly with increasing context length due to self-attention, that is not the case with UNet. We use a relatively smaller UNet for low-dimensional tasks and a UNet with a larger channel width for visual tasks. For an observation horizon of 3 and an action horizon of 15 for visuomotor tasks, the parameter count of UNet increases to (${\sim}336M$), not including the ResNet weights. UNet uses FiLM \cite{perez2018film} layers for conditioning on a single embedding, which is built for different visual inputs and proprioception similar to DP-DiT.

\textbf{FDP.} We experiment with several implementations of FDP in this paper, as shown in Figures \ref{fig:archie} $[b]$ and $[c]$. The simplest implementation shown in $[b]$ simply adds the output of the base and the residual model, with the base kept frozen. The architectures of these models are exactly identical to DP-DiT, except that they are conditioned on different modalities. For Figure \ref{fig:archie} $[c]$ used to present the results in the paper, the architecture of the base model is the same as that of DP-DiT. However, the residual model is designed similarly to the ViT \cite{dosovitskiy2020image} architecture. Since we do not denoise the inputs, we encode and pass all the inputs across the observation horizon through self-attention. We condition them on noisy actions using AdaLn. The images are encoded using patch embedding, where we keep the patch size equal to the size of the image to reduce the number of parameters. Crucially, we apply a zero layer on the block outputs of the residual model that are added to the corresponding blocks of the base model. We implement two variants of the zero-layer: a zero-initialized convolutional layer and a zero-initialized linear layer. For the convolutional layer, $\pi_{base}$ learned on proprioception is of ${\sim}30M$ parameters and the residual model $\pi_{res}$ with two camera image inputs is of ${\sim}55M$ parameters. However, the linear zero layer bloats the residual model's size to ${\sim}290M$. Other hyperparameters such as the learning rate, weight decay, and the noise schedule are the same as DP-DiT.

\subsection{Experimental Setup}
\label{app:env-setup}
\begin{figure}[!ptbh]
    \centering
    \includegraphics[width=0.8\linewidth]{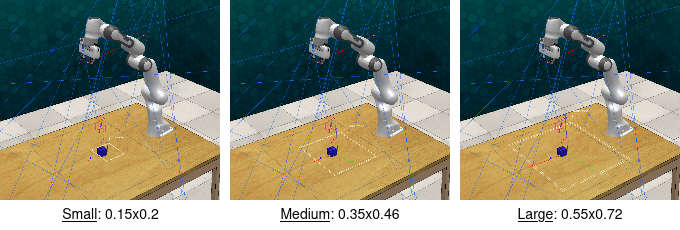}
    \caption{Task design for block pick with different scales of variation (dim in meters).}
    \label{app:fig:bp-scales}
\end{figure}

\textbf{RLBench.} We train visuomotor policies in RLBench using a multi-camera setup that records 96×96 RGB images, with joint positions as the action modality. We experiment on RLBench in both two and five camera setups (wrist, front, overhead, right-shoulder, and left-shoulder), and with an observation horizon of 2, and an action horizon of 16. Along with modifications to add distractors and color variations (Figure \ref{fig:rlbench_dist}), we also create a custom \texttt{block-pick} environment in three different sizes as shown in Figure \ref{app:fig:bp-scales}.

\begin{figure}[!ptbh]
    \centering
    \includegraphics[width=1\linewidth]{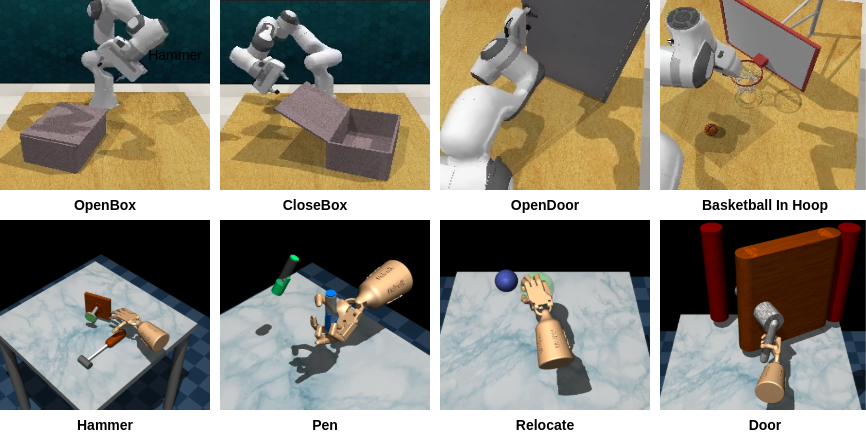}
    \caption{Selected tasks from RLBench and tasks from Adroit are shown.}
    \label{fig:task-pan}
\end{figure}

\textbf{Adroit.} The Adroit benchmark comprises high-dimensional hand manipulation tasks performed using a 24-DoF anthropomorphic hand (see Figure~\ref{fig:task-pan}). It includes four tasks—\textit{Door}, \textit{Hammer}, \textit{Pen}, and \textit{Relocate}—that demand fine motor control and complex object interaction. We modify the success condition of the \textit{Hammer} task, requiring the nail to be within a distance of 0.2 (instead of 0.1) from the board. Each Adroit task is represented by a task-specific low-dimensional state vector. We use an observation horizon of 3 and an action horizon of 15 across all Adroit experiments.

\textbf{Robomimic.} The dataset~\cite{mandlekar2021matters} provides low-dimensional state observations and uses an action space defined as the change in end-effector position and orientation (axis-angle). The benchmark includes four tasks: \textit{Lift}, \textit{Can}, \textit{Square}, and \textit{Toolhang}, with the latter two requiring higher precision. Following Chi et al. \cite{chi2023diffusion}, we use an observation horizon of 1 and an action horizon of 10 for all Robomimic experiments.

\textbf{M3L.} We use one 64$\times$64 image input and two 32$\times$32 tactile inputs to the model from the two gripper fingers holding the peg. The rotation of the end effector is frozen, and $\Delta XYZ$ is the chosen action space. Various shapes of pegs are provided by the environment, which can be inserted into the respective holes present in differently shaped blocks. In our paper, we show results for models trained on 2 pegs independently, and also on the whole assortment of pegs and blocks.

\begin{table}[t]
\centering
\scriptsize
\setlength{\tabcolsep}{2pt}
\renewcommand{\arraystretch}{1.0}
\caption{Environment specifications—Obs/Act horizon uses \texttt{oN}/\texttt{hM}. For RLBench, M3L and Real-world, env-observation depends on \#cameras (1/3/5) and 512-d embeddings.}
\label{tab:env_specs_sc}
\resizebox{\columnwidth}{!}{%
\begin{tabular}{@{}llcccccc@{}}
\toprule
\textbf{Suite} & \textbf{Task} &
\textbf{\shortstack{Env. Obs.\\Dim.}} &
\textbf{\shortstack{Rob. Obs.\\Dim.}} &
\textbf{\shortstack{Action\\Dim.}} &
\textbf{\shortstack{Max.\\Len.}} &
\textbf{\shortstack{\# Train\\Demos}} &
\textbf{\shortstack{Obs/Act\\Horizon}} \\
\midrule
\multirow{4}{*}{Robomimic}
  & Lift     & 10 & 9  & 7 & 400 & 10/50/100 & o1h10 \\
  & Can      & 14 & 9  & 7 & 400 & 10/50/100 & o1h10 \\
  & Square   & 14 & 9  & 7 & 400 & 10/50/100 & o1h10 \\
  & Toolhang & 44 & 9  & 7 & 700 & 10/50/100 & o1h10 \\
\midrule
\multirow{4}{*}{Adroit}
  & Door     & 12 & 27 & 28 & 475 & 22 & o2h16 \\
  & Hammer   & 13 & 33 & 26 & 475 & 22 & o2h16 \\
  & Pen      & 21 & 24 & 24 & 475 & 22 & o2h16 \\
  & Relocate & 9  & 30 & 30 & 475 & 22 & o2h16 \\
\midrule
RLBench    & Various   & -- & 8 & 8 & 300/600 & 10/50/100 & o3h15 \\
\midrule
M3L    & Insertion   & -- & 0 & 3 & 300 & 100/200 & o1h1 \\
\midrule
Real-World & Various   & -- & 9 & 9 & 400 & 50 & o3h15 \\
\bottomrule
\end{tabular}}
\end{table}

\textbf{Real Robot Experiments.}
The task domains used in our real-world experiments as shown in Figures \ref{fig:real-robot-exp} are described below:

\begin{figure}[!ptbh]
    \centering
    \includegraphics[width=1\linewidth]{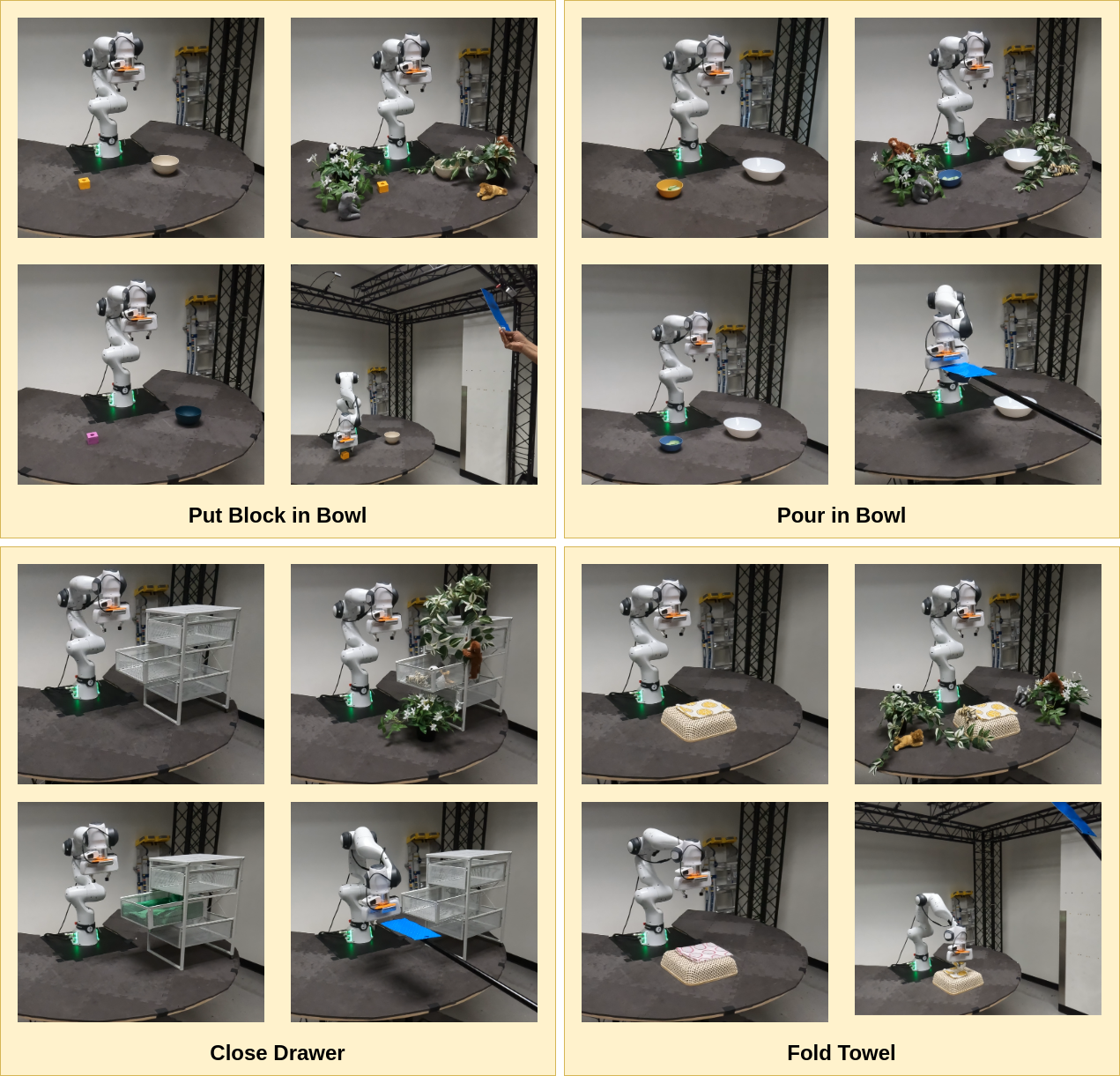}
    \caption{Tasks considered for the real robot experiments. In clockwise direction: original task, with distractors, with camera occlusions, and with color changes.}
    \label{fig:real-robot-exp}
\end{figure}

\begin{itemize}[leftmargin=*]
\item \textit{Close Drawer}: Close the cabinet of an open drawer. We vary the drawer’s placement angle and position relative to the robot within a range of $10^\circ$ and $15\,\text{cm}$, respectively. This is a relatively simple task where the robot must close the drawer by pushing it with its end-effector.
\item \textit{Put Block in Bowl}: Pick up a block and place it inside a nearby bowl. The positions of both the block and the bowl are varied within a $15\,\text{cm}$ range relative to the robot. This task assesses the policy’s ability to perform precise pick-and-place actions.
\item \textit{Pour in Bowl}: Pick up a cup and pour its contents into a nearby bowl. The positions of the cup and the bowl are varied within a $15\,\text{cm}$ range relative to the robot. This task evaluates the policy’s effectiveness in operating near joint limits.
\item \textit{Fold Towel}: Fold a kitchen towel placed on a compliant surface. The towel’s position is varied within a $5\,\text{cm}$ range relative to the robot. This task evaluates the policy’s capability in deformable object manipulation.
\end{itemize}

We used ROS1 Noetic for robot software development. For data collection, we used a 3D Connexion SpaceMouse Pro to set end-effector velocity targets, which were executed on the Franka robot using a differential inverse kinematics controller. Time-synchronized joint positions and camera images were recorded at $30$Hz for each demonstration and later post-processed by downsampling to $10$Hz for policy training. During rollout, we employed a joint position controller to sequentially execute short-horizon trajectory predictions from the policy. We allowed a trajectory length of $400$ steps for each task in all our real-world robot experiments. With a horizon length of $16$, this resulted in $25$ policy inference steps per task.

\textbf{Robot Safety Check.} We implemented a safety check in our robot software to prevent potential damage to the robot during environment variation experiments. This was particularly necessary for DP, which often generated high-jerk joint targets in out-of-distribution scenarios. For each joint command, we ensured that the target was within a threshold Euclidean distance from the current joint state, i.e., $||j_{\text{target}} - j_{\text{current}}|| \leq 0.1$. If this condition was violated, policy execution was immediately halted and the rollout was considered a failure.

\end{document}